\documentclass[10pt,twocolumn,letterpaper]{article}

\usepackage{iccv}
\usepackage{times}
\usepackage{epsfig}
\usepackage{graphicx}
\usepackage{amsmath}
\usepackage{amssymb}

\usepackage{ragged2e}
\usepackage[caption=false,font=normalsize,labelfont=sf,textfont=sf]{subfig}

\usepackage{paralist}
\usepackage{cuted}
\usepackage{multirow}
\usepackage{booktabs}
\usepackage[switch]{lineno}

\usepackage{xspace}

\usepackage{xcolor}

\newcommand*{\SCOUTERpositive}{\mbox{SCOUTER$_{+}$}\xspace}
\newcommand*{\SCOUTERnegative}{\mbox{SCOUTER$_{-}$}\xspace}

\usepackage[pagebackref=true,breaklinks=true,letterpaper=true,colorlinks,bookmarks=false]{hyperref}

\iccvfinalcopy % *** Uncomment this line for the final submission

\def\iccvPaperID{5459} % *** Enter the ICCV Paper ID here

\ificcvfinal\fi

\begin{document}

%%%%%%%%% TITLE
\title{SCOUTER: Slot Attention-based Classifier for Explainable Image Recognition}

\author{Liangzhi Li$^{1}$, Bowen Wang$^{2}$, Manisha Verma$^{1}$, Yuta Nakashima$^{1}$,\\Ryo Kawasaki$^{3}$, Hajime Nagahara$^{1}$\\
Osaka University, Japan\\
{\tt\small $^1$\{li, mverma, n-yuta, nagahara\}@ids.osaka-u.ac.jp}\\
{\tt\small $^2$bowen.wang@is.ids.osaka-u.ac.jp} {\tt\small $^3$ryo.kawasaki@ophthal.med.osaka-u.ac.jp}
}

\makeatletter
\let\@oldmaketitle\@maketitle% Store \@maketitle
\renewcommand{\@maketitle}
{%
   \newpage
   \null
   \vskip .375in
   \begin{center}
      {\Large \bf \@title \par}
      % additional two empty lines at the end of the title
      \vspace*{24pt}
      {
      \large
      \lineskip .5em
      \begin{tabular}[t]{c}
         \ificcvfinal\@author\else Anonymous ICCV submission\\
         \vspace*{1pt}\\%This space will need to be here in the final copy, so don't squeeze it out for the review copy.
Paper ID \iccvPaperID \fi
      \end{tabular}
      \par
      }
      % additional small space at the end of the author name
      \vskip .5em
      % additional empty line at the end of the title block
      \vspace*{12pt}
   \end{center}

    \vspace{-0.2in}
	\centering
	\includegraphics[width=1\textwidth]{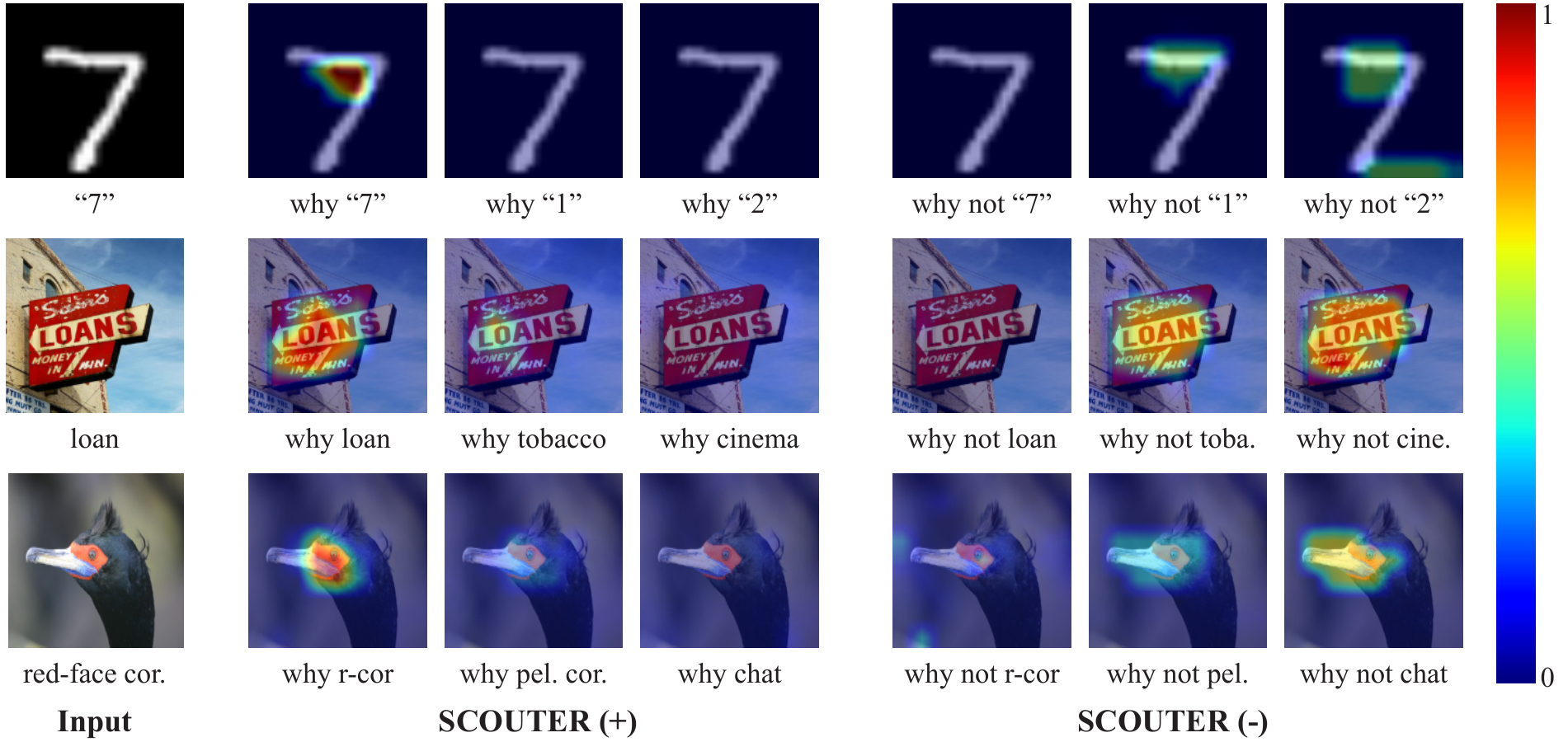}
	
	\vspace{-0.1in}
	\justify
	\refstepcounter{figure}\normalfont Figure~\thefigure. Positive and negative explanations. The images from top to down are from the test sets of MNIST \cite{MNIST}, Con-text \cite{Con-text}, and CUB-200 \cite{CUB-200} datasets. The models trained with positive ($+$) and negative ($-$) SCOUTER losses can respectively highlight the positive and negative supports, based on which one can reason why or why not the images are classified into the corresponding categories. %The attention map is used for the calculation of the weighted sum of extracted features, which will be directly used in the softmax function to output the final classification results. 
	%See Fig.~\ref{fig_supplement_raw_images} and Fig.~\ref{?} in the supplementary material for the images of the selected categories as well as the results from Grad-CAM and DeepLIFT.
	\label{fig_story}
	\vspace{4mm}
}
\makeatother

\maketitle
\ificcvfinal\thispagestyle{empty}\fi

%%%%%%%%% ABSTRACT
\begin{abstract}
\vspace{-0.1in}
Explainable artificial intelligence has been gaining attention in the past few years. However, most existing methods are based on gradients or intermediate features, which are not directly involved in the decision-making process of the classifier. In this paper, we propose a slot attention-based classifier called SCOUTER for transparent yet accurate classification. Two major differences from other attention-based methods include: (a) SCOUTER's explanation is involved in the final confidence for each category, offering more intuitive interpretation, and (b) all the categories have their corresponding positive or negative explanation, which tells ``why the image is of a certain category'' or ``why the image is not of a certain category.'' 
We design a new loss tailored for SCOUTER that controls the model's behavior to switch between positive and negative explanations, as well as the size of explanatory regions. Experimental results show that SCOUTER can give better visual explanations in terms of various metrics while keeping good accuracy on small and medium-sized datasets.
%Code is available\footnote{Code can be found in the supplementary material and will be available at GitHub after acceptance.}.
Code is available\footnote{https://github.com/wbw520/scouter}.
\end{abstract}

%%%%%%%%% BODY TEXT
\vspace{-0.23in}
\section{Introduction}\label{section_intro}
It is of great significance to know how deep models make predictions, especially for the fields like medical diagnosis, where potential risks exist when black-box models are adopted. Explainable artificial intelligence (XAI), which can give a close look into models' inference process,  % by providing  a heat map that highlights the regions that support the prediction, 
therefore has gained lots of attention.

The most popular paradigm in XAI is \textit{attributive explanation}, which gives the contribution of pixels or regions to the final prediction \cite{Grad-CAM,Ablation-CAM,IGOS,IBA}. One natural question that arises here is \textit{how these regions contribute to the decision}. %That is, attributive explanation highlights some visual patterns, of which presence/absence affects the classification, but it does \textit{not} tell whether presence of these patterns \textit{positively} affects (\ie, increases) or \textit{negatively} affects (\ie, decreases) the classification score. 
For a better view of this, %idea, we explicitly define the positive and negative patterns. Let 
let $g_{l}(v) = w_{l}^\top v + b_{l}$ denotes a fully-connected (FC) classifier for category $l$, where $w_{l}$ and $b_{l}$ are trainable vector and scalar, respectively. Training this classifier may be interpreted as a process to find from the training samples a combination of discriminative patterns $s_{{l}i}$ with corresponding weight $\gamma_i$, \ie,
\begin{equation}
    g_{l}(v) = \left(\sum_{i} \gamma_i s_{{l}i}^\top\right) v + b_{l}. 
\end{equation}
In general, these patterns can include \textit{positive} and \textit{negative} ones. Given $v$ of an image of ${l}$, a positive pattern gives $s_{{l}i}^\top v > 0$. A negative pattern, in contrast, gives $s_{{l}i}^\top v < 0$ for $v$ of any category other than ${l}$, which means that the presence of pattern described by $s_{{l}i}$ is a support of \textit{not} being category ${l}$. Therefore, set $\mathcal{S}_{l}$ of all (linearly independent) patterns for ${l}$ can be the union of sets $\mathcal{S}^+_{l}$ and $\mathcal{S}^-_{l}$ of all positive and negative patterns. 

Differentiation of positive/negative patterns gives useful information on the decision. Figure \ref{fig_story}(top) shows an MNIST image for example. One of positive patterns that makes the image being \texttt{7} can be the acute angle formed by white line segments that appears around the top-right corner, as in the second image. Meanwhile, the sixth image shows that the presence of the horizontal line is the support not being \texttt{1}. A more practical application \cite{glaucoma_xai_paper} in medical image analysis also points out the importance of visualizing positive/negative patterns. Nevertheless of the obvious benefit, recent mainstream methods like \cite{CAM,Grad-CAM,Perturbation_RISE, Perturbation_understanding} have not extensively studied this differentiation.

%The second last column of the top row in Fig.~\ref{fig_story} shows an illustrative example of such negative supports. The presence of the horizontal line of digit ``7'' is a negative support when predicting the image as digit ``1''. Although negative supports are extremely important for various applications including computer-assisted diagnosis over medical imaging, most post-hoc approaches, such as back-propagation-based ones (\eg \cite{CAM,Grad-CAM}) and perturbation-based ones \cite{Perturbation_RISE, Perturbation_understanding}, basically are ignorance of the meanings of supports since their basis lies in inferring the sensitivity or relevance between the features and the decision confidences. The attention-based approach (\eg \cite{kim2017interpretable,Google_attention_xai_ocr}) cannot distinguish positive/negative supports because black-box classification layers after the explanatory layer apply nonlinear transformation to the features. 

Positive and negative patterns lead to two interesting questions: (i) Can we provide \textit{positive explanation} and \textit{negative explanation} that visually show support regions in the image that correspond to positive and negative patterns? (ii) As the combination of patterns to be learned is rather arbitrary and any combination is possible as long as it is discriminative; can we provide preference on the combination in order to leverage prior knowledge on the task in training? 

In this paper, we re-formulate explainable AI with an \textit{explainable classifier}, coined SCOUTER (Slot-based COnfigUrable and Transparent classifiER), which tries to find either positive or negative patterns in images. This approach is similar to the attention-based approach (\eg \cite{kim2017interpretable,Google_attention_xai_ocr}) rather than the post-hoc approaches \cite{CAM, Grad-CAM, IBA}. 
%
%Supposing that, for each category $l$, there exists a \textit{support set} $\mathcal{S}_l=\{s_{l1}, s_{l2}, \dots\}$, in which the elements are in support of the decision towards/against category $l$ for the input image. SCOUTER is designed to find a subset $\mathcal{S}'_l \in \mathcal{S}_l$ that includes one or several supports from $\mathcal{S}_l$. The decisions by SCOUTER are solely based on the presence of $\mathcal{S}=\{\mathcal{S}'_l|l=1,2,\dots\}$ found in the image, without using a black-box classifier that makes $\mathcal{S}$ less interpretable. This transparency enables SCOUTER configurable to find either positive ($\mathcal{S}_+$) or negative ($\mathcal{S}_-$) supports, of which visualization can serve as positive or negative explanations. 
%
%With this new paradigm of \textit{explainable classifiers}, smaller support regions may be preferable to facilitate the semantic interpretation of each support. That is, finding a combination of eyes, noses, lips, \etc~may offer more explainability than directly finding a face. We thus introduce a new term in the loss function to constrain the size of support sizes. Such a constraint over the support sizes may deteriorate the classification performance by itself, but combinations of multiple supports can compensate for missing clues. 
%
Our newly proposed explainable slot attention (xSlot) module is the main building block of SCOUTER. This module is built on top of the recently-emerged slot attention \cite{Slot}, which offers an object-centric approach for image representation. The xSlot module identifies the spatial support of either positive or negative patterns for each category in the image, which is directly used as the confidence value of that category; the commonly-used FC classifier is no longer necessary. The xSlot module can also be used to visualize the support as shown in Fig.~\ref{fig_story}. SCOUTER is also characterized by its configurability over patterns to be learned, \ie, the choice of positive or negative pattern and the desirable size of the pattern, which can incorporate the prior knowledge on the task. The controllable size of explanation can be beneficial for some applications, \eg, disease diagnosis in medicine, defect recognition in manufacturing, \etc.   

\vspace{-0.23in}
\paragraph{Contribution}
Our transparent classifier, SCOUTER, explicitly models positive and negative patterns with a dedicated loss, allowing to set preference over the spatial size of patterns to be learned. We experimentally show that SCOUTER successfully learns both positive and negative patterns and visualize their support in the given image as the explanation, achieving state-of-the-art results in several commonly-used metrics like IAUC/DAUC \cite{Perturbation_RISE}. Our case study in medicine also highlights the importance of both types of explanations as well as controlling the area size of explanatory regions.

%The main contributions of our work include:
%\begin{compactitem}
%	\item We propose a transparent classifier that gives precise and meaningful explanations, which are directly involved in the decision-making process.
%	\item We design a loss to adjust the exploratory regions for different tasks/datasets, which can better meet the requirements of various applications. 
%	\item We introduce novel concepts of positive and negative explanations, of which the latter one is a new type of explanation that can be extremely helpful when machine decisions are against users' expectations.
%\end{compactitem}

\section{Related Work}

\subsection{Explainable AI} \label{section-related-work-XAI}

There are mainly three XAI paradigms \cite{survey_explainable}, \ie \textit{post-hoc}, \textit{intrinsic}, and \textit{distillation}.
The post-hoc paradigm usually provides a heat map highlighting important regions for the decision (\eg \cite{Grad-CAM,IBA}). The heat map is computed besides the forward path of the model. The intrinsic paradigm explores the important piece of information within the forward path of the model, \eg, as attention maps (\eg \cite{kim2017interpretable, Google_attention_xai_ocr, mascharka2018transparency, xie2019visual}). \textit{Distillation methods} are built upon model distillation \cite{distilling}. The basic idea is to use an inherently transparent model to mimic the behaviors of a black-box model (\eg \cite{ zhang2019interpreting, Distillation_LIME}). 

The post-hoc paradigm has been extensively studied among them. The most popular type of methods is based on channel activation or back-propagation, including CAM \cite{CAM}, GradCAM \cite{Grad-CAM}, DeepLIFT \cite{DeepLIFT}, and their extensions \cite{GradCAM++, Smooth_GradCAM++, Score-CAM, SS-CAM, Ablation-CAM}.  Another type of method is perturbation-based, including Occlusion \cite{Perturbation_visualizingCNN}, RISE \cite{Perturbation_RISE}, meaningful perturbations \cite{Perturbation_Meaningful}, real-time saliency \cite{Perturbation_realtime}, extremal perturbations \cite{Perturbation_understanding}, I-GOS \cite{IGOS}, IBA \cite{IBA}, \etc. These methods basically give \textit{attributive explanation}, which visualizes support regions of learned patterns for each category ${l}$ in the set of all possible categories $\mathcal{L}$. This visualization can be done by finding regions in feature maps or the input image that give large impact on the score $g_{l}$. By definition, attributive explanation is the same as our positive explanation. 

Some of the methods for attributive explanation thus can be extended to provide negative explanations by negating the sign of the score, feature maps, or gradients. %\textcolor{blue}{However, this implementation may not be enough to pinpoint negative patterns in the image because, for FC classifiers with softmax cross-entropy, negative patterns are learned only through the denominator of softmax, which is the sum of all scores}. 
It should be noted that the interpretation of visual explanation by gradient-based methods \cite{Grad-CAM, GradCAM++, Smooth_GradCAM++} may not be straightforward because of linearization of $g_{l}$ for the given image; and thus the resulting visualization may not highlight the support regions for negative patterns. GradCAM \cite{Grad-CAM} refers to its negative variant as \textit{counterfactual explanation} that gives regions that can change the decision, emphasizing how it should be interpreted. 

\textit{Discriminant explanation} is a new type of XAI in the post-hoc paradigm, which appeared in \cite{Scout} to show ``why image $x$ belongs to category ${l}$ rather than ${l}'$.'' This can be interpreted using set $\mathcal{S}_{l}^+$ of all possible positive patterns for ${l}$ and set $\mathcal{S}_{{l}'}^-$ of all possible negative patterns for ${l}'$: It try to spot a (combination of) discriminative pattern $s$ that is in the intersection $\mathcal{S}_{l}^+ \cap \mathcal{S}_{l'}^-$. Due to the unavailability of negative patterns, the method \cite{Scout} first finds (a combination of) positive patterns and uses the complementary of the region containing the positive patterns as a proxy of negative patterns. Goyal \etal gives another line of counterfactual explanation in \cite{goyal2019counterfactual}. Given two images of categories $l$ and $l'$, they find the region in the image of $l$, of which replacement to a certain region in the image of $l'$ changes the prediction from $l$ to $l'$. This can be also implemented using discriminant explanation.

SCOUTER computes a heat map to spot regions important for the decision in the forward path, so it falls into the intrinsic paradigm. Together with the dedicated loss, it can directly identify positive and negative patterns with control over the size of patterns.

\subsection{Self-attention in Computer Vision}

Self-attention is first introduced in the Transformers \cite{Transformer}, in which self-attention layers scan through the input elements one by one and update them using the aggregation over the whole input. Initially, self-attention is used in place of recurrent neural networks for sequential data, \eg, natural language processing \cite{BERT}. Recently, self-attention is adopted to the computer vision field, \eg, Image Transformer \cite{ImageTransformer}, Axial-DeepLab \cite{Axial-DeepLab}, DEtection TRansformer (DETR) \cite{DETR}, Image Generative Pre-trained Transformer (Image GPT) \cite{Image_GPT}, \etc. Slot attention \cite{Slot} is also based on this mechanism to extract object-centric features from images (there are some other works \cite{henaff2016tracking, goyal2020object} using the concept of \textit{slot}); however, the original slot attention is tested only on some synthetic image datasets. SCOUTER is based on slot attention but is designed to be an explainable classifier applicable to natural images.

\begin{figure}[!t]
	\centering
	\subfloat[]{\includegraphics[width=.9\columnwidth]{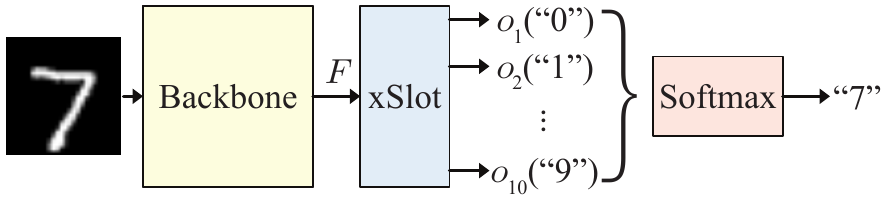}}
	\vspace{-0.1in}
	\subfloat[]{\includegraphics[width=.9\columnwidth]{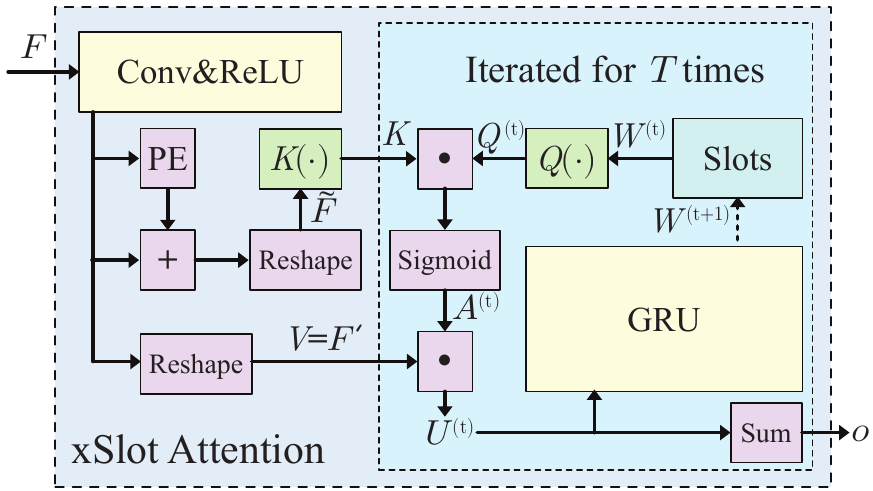}}
	\caption{Classification pipeline. (a) Overview of the classification model. (b) The xSlot Attention module in SCOUTER.}
	\label{fig_structure}
\end{figure}

\section{SCOUTER}

Given an image $x$, the objective of a classification model is to find its most probable category $l$ in category set $\mathcal{L}=\{l_1, l_2,\dots,l_n\}$. This can be done by first extracting features $F = B(x) \in\mathbb{R}^{c \times h\times w}$ using a backbone network $B$.
$F$ is then mapped into a score vector $o \in\mathbb{R}^{n}$, representing the confidence values, using FC layers and softmax as the classifier. However, such FC classifiers can learn an arbitrary (nonlinear) transformation and thus can be black-box.  %some capacity.

We replace such an FC classifier with our SCOUTER (Fig.~\ref{fig_structure}(a)), consisting of the xSlot attention module, which produces the confidence for each category given features $F$. The whole network, including the backbone, is trained with the \textit{SCOUTER loss}, which provides control over the size of support regions and switching between positive and negative explanations.% We will introduce these two concepts in the following subsections.

\subsection{xSlot Attention}
%To implement an explainable classifier, we have modified several important designs in the original slot attention mechanism \cite{Slot}. Four major differences are listed below. (i) Slots in our paper are allocated for each object category, rather than a neck module connecting the backbone and downstream networks. %We define that all categories have their own slot modules. 
%(ii) To better differentiate the categories, we make the weights of slots fully learnable, which is different from the original slot attention module, in which the slot weights are always randomly generated and only the parameters of the random function are trainable. (iii) There is no softmax function in the dimension of slots, therefore, slots do not compete with each other, which is the key design in its original slot attention. (iv) We remove the multilayer perceptron (MLP) module as well as the linear transformation for $\mathcal{V}$ and changed the output of slot attention to make it more transparent.

In the original slot attention mechanism \cite{Slot}, a \textit{slot} is a representation of a local region aggregated based on the attention over the feature maps. A single slot attention module with multiple slots is attached on top of the backbone network $B$. Each slot produces its own feature as output. This configuration is handy when there are multiple objects of interest. This idea can be transferred to spot the supports in the input image that leads to a certain decision. 

The main building block of SCOUTER is the xSlot attention module, which is a variant of the slot attention module tailored for SCOUTER. Each slot of the xSlot attention module is associated with a category and gives the confidence that the input image falls into the category. With the slot attention mechanism, the slot for category $l$ is required to find support $\mathcal{S}_l$  in the image that directly correlates to $l$.

Given feature $F$, the xSlot attention module updates slot $w^{(t)}_l$ for $T$ times, where $w^{(t)}_l$ represents the slot after $t$ updates and $l \in \mathcal{L}$ is the category associated to this slot. The slot is initialized with random weights, \ie,
\begin{equation}
    w^{(0)}_l \sim \mathcal{N}(\mu, \text{diag}(\sigma)) \in \mathbb{R}^{1 \times c'},
\end{equation}
where $\mu$ and $\sigma$ are the mean and variance of a Gaussian, and $c'$ is the size of the weight vector. We denote the slots for all categories by $W^{(t)} \in \mathbb{R}^{n \times c'}$.

The slot $W^{(t+1)}$ is updated using $W^{(t)}$ and feature $F$. Firstly, $F$ goes through a $1\times 1$ convolutional layer to reduce the number of channels and the ReLU nonlinearity as $F' = \operatorname{ReLU}(\operatorname{Conv}(F)) \in \mathbb{R}_{+}^{c' \times d}$, with $F$'s spatial dimensions being flattened ($d = hw$). $F'$ is augmented by adding the position embedding to take the spatial information into account, following \cite{Transformer, DETR}, \ie $\tilde{F} = F' + \text{PE}$, where PE is the position embedding. We then use two multilayer perceptrons (MLPs) $Q$ and $K$, each of which has three FC layers and the ReLU nonlinearity between them. This design is for giving more flexibility in the computation of \textit{query} and \textit{key} in the self-attention mechanism. Using
\begin{equation}
    Q(W^{(t)}) \in \mathbb{R}^{n \times c'}, \quad K(\tilde{F}) \in \mathbb{R}^{c' \times d},
\end{equation}
we obtain the dot-product attention $A^{(t)}$ using sigmoid $\sigma$ as
\begin{equation} \label{eq_QxK}
    A^{(t)} = \sigma(Q(W^{(t)}) K(\tilde{F})) \quad \in (0,1)^{n \times d}.
\end{equation}

%The reason we choose dot-product, which is actually matrix multiplication, rather than additive attention, which is computed by an extra hidden layer, is because the former one is much faster and space-efficient in practice due to the code optimization \cite{Transformer}.

The attention is used to compute the weighted sum of features in the spatial dimensions by
\begin{equation}\label{eq:feature}
    U^{(t)} = A^{(t)} {F'}^\top \quad \in \mathbb{R}^{n \times c'},
\end{equation}
and slot $W^{(t)}$ is eventually updated through a gated recurrent unit (GRU) as
\begin{equation} \label{eq_GRU}
W^{(t+1)}=\operatorname{GRU}(U^{(t)}, W^{(t)}),
\end{equation}
taking $U^{(t)}$ and $W^{(t)}$ as input and hidden state, respectively. Following the original slot attention module, we update the slot for $T = 3$ times. 

The output of the xSlot attention module is the sum of all elements for category $l$ in $U^{(T)}$, which is a function of $F$. Formally, the output of xSlot Attention module is:
\begin{equation}\label{eq:cls}
\text{xSlot}(F) = U^{(T)} \mathbf{1}_{c'}  \quad \in \mathbb{R}_+^n,
\end{equation}
where $\mathbf{1}$ is the column vector with all $c'$ elements being 1. 

From Eqs.~(\ref{eq:feature}) and (\ref{eq:cls}), we have $\text{xSlot}(F)=A^{(T)}F'^\top\mathbf{1}_{c'}$, where $F'^\top\mathbf{1}_{c'} \in \mathbb{R}^{d}$ is a reduction of $F$ and is a class-agnostic map. The $l$-th row of $A^{(T)}$ can then be viewed as spatial weights over map $F'^\top\mathbf{1}_{c'}$ to spot where the support regions for category $l$ is\footnote{$F'^\top\mathbf{1}_{c'}$ can be viewed as a single map that includes a mixture of supports for all categories.}. In order for visualizing the support regions, we reshape and resize each row of $A^{(T)}$ to the input image size.

Note that in the original slot attention module, a linear transformation is applied to the features, \ie, $V(\tilde{F})$, which is then weighted using Eq.~(\ref{eq:feature}). However, the xSlot attention module omits this transformation as it already has a sufficient number of learnable parameters in $Q$, $K$, GRU, \etc., and thus the flexibility. Also, the confidences, given by Eq.~(\ref{eq:cls}), are typically computed by an FC layer, while SCOUTER just sums up the output of xSlot attention module, which is actually the presence of learned supports for each category. This simplicity is essential for a transparent classifier as discussed in Section \ref{sec:switch}.

\subsection{SCOUTER Loss}

The whole model, including the backbone network, can be trained by simply applying softmax to xSlot$(F)$ and minimizing cross-entropy loss $\ell_\text{CE}$. However, there is a phenomenon that, in some cases, the model prefers attending to a broad area (\eg, a slot covers a combination of several supports that occupy large areas in the image) depending on the content of the image. As argued in Section \ref{section_intro}, it can be beneficial to have control over the area of support regions to constrain the coverage of a single slot. 

Therefore, we design the SCOUTER loss to limit the area of support regions. The SCOUTER loss is defined by
\begin{equation} \label{eq:loss}
\ell_{\operatorname{SCOUTER}} = \ell_{\operatorname{CE}}+ \lambda \ell_{\operatorname{Area}},
\end{equation}
where $\ell_{\operatorname{Area}}$ is the area loss, $\lambda$ is a hyper-parameter to adjust the importance of the area loss. The area loss is defined by
\begin{equation}
\ell_{\operatorname{Area}} = \mathbf{1}_n^\top A^{(T)} \mathbf{1}_d,
\end{equation}
which simply sums up all the elements in $A^{(T)}$. With larger $\lambda$, SCOUTER attends smaller regions by selecting fewer and smaller supports. On the contrary, it prefers a larger area with smaller $\lambda$. %We will show in the experiment section the relationship between $\lambda$ and area sizes as well as the classification accuracy. 

%where $p$ is a factor to adjust the convergence speed of area loss. With a larger $p$, area loss will have a smaller gradient when $L_{\operatorname{Area}}$ is close to $0$. Therefore, by setting a larger $p$, we can move the focus of the model training from minimizing the area size to improving the classification accuracy if the area size is already small enough. More details are also present in the ablation study.

\subsection{Switching Positive and Negative Explanation}
\label{sec:switch}

The model with the SCOUTER loss in Eq.~(\ref{eq:loss}) can only provide positive explanation since larger elements in $A^{(T)}$ means the prediction is made based on the corresponding features. We introduce a hyper-parameter $e \in \{+1, -1\}$ in Eq.~(\ref{eq:cls}), \ie,
\begin{equation}\label{eq:posneg}
o = \text{xSlot}_e(F) = e \cdot U^{(T)} \mathbf{1}_{c'}  \quad \in \mathbb{R}_+^n,
\end{equation}
where $o=\{o_1,\dots,o_n\}$. This hyper-parameter configures the xSlot attention module to learn to find either positive or negative supports. 

%Since $\mathcal{U}$'s elements are all in $\mathbb{R}_+$, the entire model with the xSlot Attention module will have different behaviors for $e$ being either $1$ or $-1$. 

With the softmax cross-entropy loss, the model learns to give the largest confidence $o_l$ corresponding to ground-truth (GT) category $l$ and a smaller value $o_{l'}$ to wrong category $l' \neq l$.
For $e=+1$, all elements given by $\operatorname{xSlot}$ is non-negative since both $A^{(T)}$ and $F'$ are non-negative and thus $U^{(T)}$ is. For arbitrary non-negative $F'$, thanks to simple reduction in Eq.~(\ref{eq:cls}), larger $o_l$ can be produced only when some elements in $a^{(T)}_l$, the row vector in $A^{(T)}$ corresponding to $l$, is close to 1, whereas a smaller $o_{l'}$ is given when all elements in $a^{(T)}_l$ are close to 0. Therefore, by setting $e$ to $+1$, the model learns to find the positive supports $\mathcal{S}^{+}_{l}$ among the images of the GT category. The visualization of $a^{(T)}_l$ thus serves as positive explanation, as in Fig.~\ref{fig_story} (left).

On the contrary, for $e = -1$, all elements in $o$ are negative and thus the prediction by Eq.~(\ref{eq:posneg}) gives $o_l$ close to 0 for correct category $l$ and smaller $o_{l'}$ for non-GT category $l'$. To make $o_l$ close to 0, all elements in $a^{(T)}_l$ must be close to 0, and a smaller $o_{l'}$ is given when $a^{(T)}_{l'}$ has some elements close to 1. For this, the model learns to find the negative supports $\mathcal{S}_-$ that do not appear in the images of the GT category. As a result,  $a^{(T)}_{l'}$ can be used as negative explanation, as shown in Fig.~\ref{fig_story} (right).

\section{Experiments}\label{section_experiments}

\subsection{Experimental Setup}
We chose to use the ImageNet dataset \cite{ImageNet} for a detailed evaluation of SCOUTER, because of the following three reasons: (i) It is commonly used in the evaluation of classification models. (ii) There are many classes with similar semantics and appearances, and the relationships among them are available in the synsets of the WordNet, which can be used to evaluate positive and negative explanations. (iii) Bounding boxes are available for foreground objects, which helps measure the accuracy of visual explanation. In experiments, we use subsets of ImageNet by extracting the first $n$ ($0 < n \leq 1,000$) categories in the ascending order of the category IDs. Also, we present classification performance on Con-text \cite{Con-text} and CUB-200 \cite{CUB-200} datasets and illustrate glaucoma diagnosis using quantitative and qualitative results on ACRIMA \cite{ACRIMA} dataset. 

The size of images is $260\times 260$. 
%Our models are implemented with PyTorch. 
%AdamW \cite{AdamW} %, a variant of the original Adam algorithm \cite{Adam}, 
%is adopted as the optimizer with the initial learning rate of $10^{-4}$. %The implementations and pre-trained weights of the underlying CNN models are extracted from the \textit{PyTorch Image Models (timm)} library. We adopt the \textit{imgaug} library for the data augmentation. 
The feature $F$ extracted by the backbone network is mapped into a new feature $F'$ with the channel number $c'=64$.
%All the experiments are conducted on three local GPU servers 
%For efficiency, we use three servers at the same time to run experiments in parallel. These servers are with same hardware (GPU\&CPU). % and with a shared network file system in which data are synchronized in real-time. 
%equipped with two Intel Xeon Gold 5122 (@3.60GHz) CPUs, four NVIDIA Tesla V100-SXM2 (32GB) GPUs, and 384GB memory. Every run of training occupied one single GPU.% No other processes will utilize that GPU as well as its spare memory to avoid possible interference.
The models were trained on the training set for $20$ epochs and the performance scores are computed on the validation set with the trained models after the last epoch. All the quantitative results are obtained by averaging the scores from three independent runs.

\begin{table*}[!t]
	\caption{Evaluation of the explanations. Positive explanations are from the GT class, while negative is from the least similar class (LSC).}
	\label{table_exp_explainability_measurement}
	\centering
	\resizebox{0.975\textwidth}{!}{%
    	\begin{tabular}{clcccccccc}
    		\toprule
    	    & Methods & Year & Type & Area Size & Precision $\uparrow$ & IAUC $\uparrow$ & DAUC $\downarrow$ & Infidelity $\downarrow$ & Sensitivity $\downarrow$ \\
    	    \midrule
            \multirow{14}{*}{Positive} & CAM \cite{CAM} & 2016 & Back-Prop & 0.0835 & 0.7751 & 0.7185 & 0.5014 & 0.1037 & 0.1123 \\
            & GradCAM \cite{Grad-CAM} & 2017 & Back-Prop & 0.0838 & 0.7758 & 0.7187 & 0.5015 & 0.1038 & 0.0739\\
            %& DeepLIFT \cite{DeepLIFT} & 2017 & Back-Prop & 0.0874 & 0.7504 & 0.7207 & 0.4699 & 0.1042 & 0.0800\\
            & GradCAM++ \cite{GradCAM++} & 2018 & Back-Prop & 0.0836 & 0.7861 & 0.7306 & 0.4779 & 0.1036 & 0.0807 \\
            & S-GradCAM++ \cite{Smooth_GradCAM++} & 2019 & Back-Prop & 0.0868 & 0.7983 & 0.6991 & 0.4896 & 0.1548 & 0.0812\\
            & Score-CAM \cite{Score-CAM} & 2020 & Back-Prop & 0.0818 & 0.7714 & 0.7213 & 0.5247 & 0.1035 & 0.0900\\
            & SS-CAM \cite{SS-CAM} & 2020 & Back-Prop & 0.1062 & 0.7902 & 0.7143 & 0.4570 & 0.1109 & 0.1183\\
            & $\quad \llcorner$ w/ threshold & 2020 & Back-Prop & 0.0496 & 0.8243 & 0.6010 & 0.7781 & 0.1079 & 0.0790 \\
            & RISE \cite{Perturbation_RISE} & 2018 & Perturbation & 0.3346 & 0.5566 & 0.6913 & 0.4903 & 0.1199 & 0.1548\\
            & Extremal Perturbation \cite{Perturbation_understanding} & 2019 & Perturbation & 0.1458 & 0.8944 & 0.7121 & 0.5213 & 0.1042 & 0.2097\\
            & I-GOS \cite{IGOS} & 2020 & Perturbation & 0.0505 & 0.8471 & 0.6838 & 0.3019 & 0.1106 & 0.6099\\
            & IBA \cite{IBA} & 2020 & Perturbation & 0.0609 & 0.8019 & 0.6688 & 0.5044 & 0.1039 & 0.0894\\
            & \SCOUTERpositive ($\lambda=1$) &  &Intrinsic& 0.1561 & 0.8493 & 0.7512 & 0.1753 & \textbf{0.0799} & 0.0796\\
            & \SCOUTERpositive ($\lambda=3$) &  &Intrinsic& 0.0723  & 0.8488 & \textbf{0.7650} & \textbf{0.1423} & 0.0949 & \textbf{0.0608}\\
            & \SCOUTERpositive ($\lambda=10$) &  &Intrinsic& 0.0476 & \textbf{0.9257} & 0.7647 & 0.2713 & 0.0840 & 0.1150\\
            \midrule
            \multirow{9}{*}{Negative} & CAM \cite{CAM} & 2016 & Back-Prop & 0.1876 & 0.3838 & 0.6069 & 0.6584 & 0.1070 & 0.0617 \\
            & GradCAM \cite{Grad-CAM} & 2017 & Back-Prop & 0.0988 & 0.6543 & 0.6289 & 0.7281 & 0.1060 & 0.5493 \\
             & GradCAM++ \cite{GradCAM++} & 2018 & Back-Prop & 0.0879 & 0.6280 & 0.6163 & 0.6017 & 0.1047 & 0.3114 \\
            & S-GradCAM++ \cite{Smooth_GradCAM++} & 2019 & Back-Prop & 0.1123 & 0.6477 & 0.6036 & 0.5430 & 0.1071 & 0.0590\\
            & RISE \cite{Perturbation_RISE} & 2018 & Perturbation & 0.4589 & 0.4490 & 0.4504 & 0.7078 & 0.1064 & 0.0607\\
            & Extremal Perturbation \cite{Perturbation_understanding} & 2019 & Perturbation & 0.1468 & 0.6390 & 0.2089 & 0.7626 & 0.1068 & 0.8733 \\
            & \SCOUTERnegative ($\lambda=1$) &  &Intrinsic& 0.0643 & 0.8238 & \textbf{0.7343} & \textbf{0.1969} & 0.0046 & \textbf{0.0567}\\
            & \SCOUTERnegative ($\lambda=3$) &  &Intrinsic& 0.0545  & \textbf{0.8937} & 0.6958 & 0.4286 & 0.0196 & 0.1497 \\
            & \SCOUTERnegative ($\lambda=10$) &  &Intrinsic&  0.0217 & 0.8101 & 0.6730 & 0.7333 & \textbf{0.0014} & 0.1895\\
    		\bottomrule
    	\end{tabular}
	}
\end{table*}

\begin{table}[!t]
	\caption{Area sizes of the explanations ($\lambda=10$).}
	\label{table_exp_pos_neg_compare}
	\centering
	\resizebox{0.95\columnwidth}{!}{%
    	\begin{tabular}{ccccc}
    		\toprule
    	    \multirow{2}{*}{Methods} & \multicolumn{4}{c}{Target Classes}\\\cline{2-5}
    	    &GT  &  Highly-similar & Similar  & Dissimilar \\
    	    \midrule
            \SCOUTERpositive & 0.0476 & 0.0259 & 0.0093 & 0.0039\\
            \SCOUTERnegative & 0.0097 & 0.0141 & 0.0185 & 0.0204\\
    		\bottomrule
    	\end{tabular}
	}
	
\end{table}

\subsection{Explainability}

%Explainability is usually evaluated qualitatively or with some simple quantitative tests from machine teaching experiments \cite{goyal2019counterfactual}, which are subjective and may not be so convincing \cite{Scout}. Recently, a new evaluation metric using the intersection over union (IoU) between explanations and semantic masks is proposed \cite{Scout}. However, this is designed for the counterfactual explanation, in which two categories are involved in one explanation. Also, the proposed method prefer to generating small explanations when $\lambda$ is large, rather than having the same shape and size as the objects of interest. Therefore, the intersection between our explanation and the mask can be always small and thus IoUs are biased toward 0. 

%We instead evaluate the accuracy of our visual explanation by the precision that measures how much regions spotted by our explanation are covered by the objects of interest. 
To evaluate the quality of visual explanation, we use bounding boxes provided in ImageNet as a proxy of the object regions and compute the percentage of the pixels located inside the bounding box over the total pixel numbers in the whole explanation. Specifically, for set $I$ of all pixels in the input image and set $D$ of all pixels in the bounding box, we define the explanation precision as $\operatorname{Precision}_l = \frac{ \sum_{i \in D} a^l_i}{\sum_{i \in I} a^l_i}$,
where category $l \in \mathcal{L}$ and $a^l_i \in [0,1]$ is the value of pixel $i$ in $\bar{A_l}$, which is attention map $A$ resized to the same size as the input image by bilinear interpolation. %This metric is associated with category $l$ since SCOUTER produces different support regions for different categories. For the positive explanation, the GT category's precision must be close to 1, while the other categories' precision is not necessarily high. 
We compute this metric on the visualization results of the GT category for positive explanations and on the least similar class (LSC) (using Eq.~\ref{eq_wordnet_similarity}) for negative explanations, as LSC images usually show strong and consistent negative explanations.
This precision metric actually is a generalization of the pointing game \cite{top_down}, which counts one \textit{hit} when the point with the largest value on the heat map locates in the bounding box and the final score is calculated as $\frac{\operatorname{\#Hits}}{\operatorname{\#Hits}+\operatorname{\#Misses}}$.

We also adopt several other metrics, \ie, 
(i) insertion area under curve (IAUC) \cite{Perturbation_RISE}, which measures the accuracy gain of a model when gradually adding pixels according to their importance given in the explanation (heat map) to a synthesized input image; %As a sharp performance increase is expected for the good explanations, therefore, the higher IAUC the better. 
(ii) deletion area under curve (DAUC) \cite{Perturbation_RISE}, which measures the performance drop when gradually removing important pixels from the input image; %Similarly, we would like to see that good explanations can lead to significant drops, thus, the lower DAUCs the better.
(iii) infidelity \cite{infidelity_sensitivity}, which measures the degree to which the explanation captures how the prediction changes in response to input perturbations; and 
(iv) sensitivity \cite{infidelity_sensitivity}, which measures the degree to which the explanation is affected by the input perturbations.
In addition, we calculate the (v) overall size of the explanation areas by $\operatorname{Area}_l = \sum_{i \in I} a^l_i$,
as for some applications, a smaller value is better to pinpoint the supports to differentiate one class from the others. 

We conduct the explainability experiments with the ImageNet subset with the first 100 classes. We train seven models with (1) an FC classifier, (2)--(4) \SCOUTERpositive ($\lambda=1,3,10$), and (5)--(7) \SCOUTERnegative ($\lambda=1,3,10$) using ResNeSt 26 \cite{ResNeSt} as the backbone. The results of competing methods are obtained from the FC classifier-based model.
In addition, as introduced in Section~\ref{section-related-work-XAI}, some of the existing works can give negative explanations. Therefore, we also implement and compare our results with their negative variants by using negative feature maps/gradients or modifying their objective functions.

The numerical results are shown in Table \ref{table_exp_explainability_measurement}. We can see that SCOUTER can generate explanations with different area sizes while achieving good scores in all metrics. These results demonstrate that the visualization by SCOUTER is preferable in terms of controlling area sizes, high precision, insensitive to noises (sensitivity), and with good explainability (infidelity, IAUC, and DAUC).
%Note that this test is performed on the images of ground-truth (GT) class, therefore, \SCOUTERnegative gives less explanations than \SCOUTERpositive. 
%In addition, negative explanations on LSC does not perform well on IAUC and DAUC. This is because these two metrics are measured on GT, and the supports to deny LSC do not necessarily have the essential information to admit GT.
%In addition, \SCOUTERpositive shows higher precision on non-GT classes than the GT class, while \SCOUTERnegative performs oppositely. This is mainly because the background pixels get fewer inferences from the foreground pixels when the attentions on foreground pixels get weaker. Therefore, the overall precision becomes higher.
Among the competing methods, extremal perturbation \cite{Perturbation_understanding}, I-GOS \cite{IGOS}, and IBA \cite{IBA} also take the size of support regions into account, and thus some of them give smaller explanatory regions. Extremal perturbation's explanatory regions cover some parts of foreground objects. This leads to a high precision score, but the performance over other metrics is not satisfactory. I-GOS and IBA give small explanation areas. I-GOS results in low IAUC and sensitivity scores. IBA gets relatively low scores of IAUC and DAUC, which means its explanations cannot give correct attention to the pixels and thus does not have enough explainability.

It is arguable that area size can be controlled by thresholding the heatmap. In order to verify this, we set a threshold ($a\geq 0.2$) to one of the back-propagation-based methods (SS-CAM) to get explanations with smaller size. We can see that this variant suffers a deterioration in IAUC and DAUC (significantly worse than I-GOS, IBA, and SCOUTER), which represents a large explainability drop and hampers its uses in actual applications requiring small explanations.

To further explore the explanation for non-GT categories, we define the semantic similarity between categories based on \cite{Wu_Palmer}, which uses WordNet, as
\begin{equation} \label{eq_wordnet_similarity}
    \operatorname{Similarity} = 2\frac{d(\operatorname{LCS}(l,l'))}{d(l)+d(l')},
\end{equation}
where $d(\cdot)$ gives the depth of category $l$ in WordNet, and $\operatorname{LCS}(l,l')$ is to find the least common subsumer of two arbitrary categories $l$ and $l'$. We define the highly-similar categories as the category pairs with a similarity score no less than $0.9$, similar categories as with a score in $[0.7, 0.9)$, and the remaining categories are regarded as dissimilar categories.
Table \ref{table_exp_pos_neg_compare} summarizes the area sizes of the explanatory regions for GT, highly-similar, similar, and dissimilar categories. We see a clear trend: \SCOUTERpositive decreases the area size when the inter-category similarity becomes lower, while \SCOUTERnegative %shows the opposite trend. 
gives larger explanatory regions for the dissimilar categories.

\begin{figure*}[!t]
	\centering
	\includegraphics[width=\textwidth]{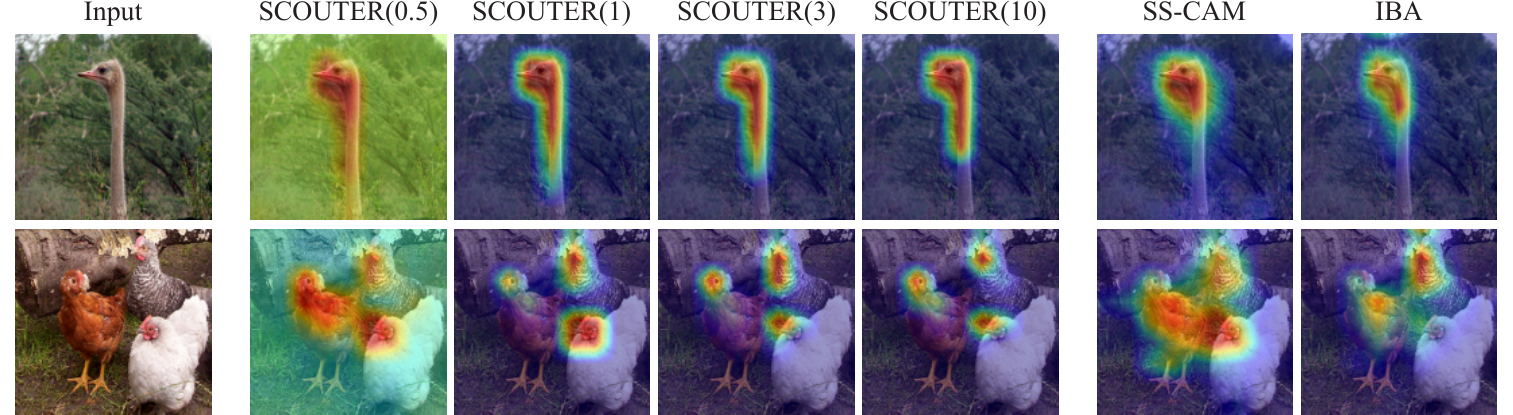}
	\caption{Visualized positive explanations using \SCOUTERpositive and existing methods. The numbers in the parentheses are the $\lambda$ values used in the SCOUTER training.}
	\vspace{-0.1in}
	\label{fig_exp_area_size_examples}
\end{figure*}

\begin{figure}[!t]
	\centering
	\includegraphics[width=\columnwidth]{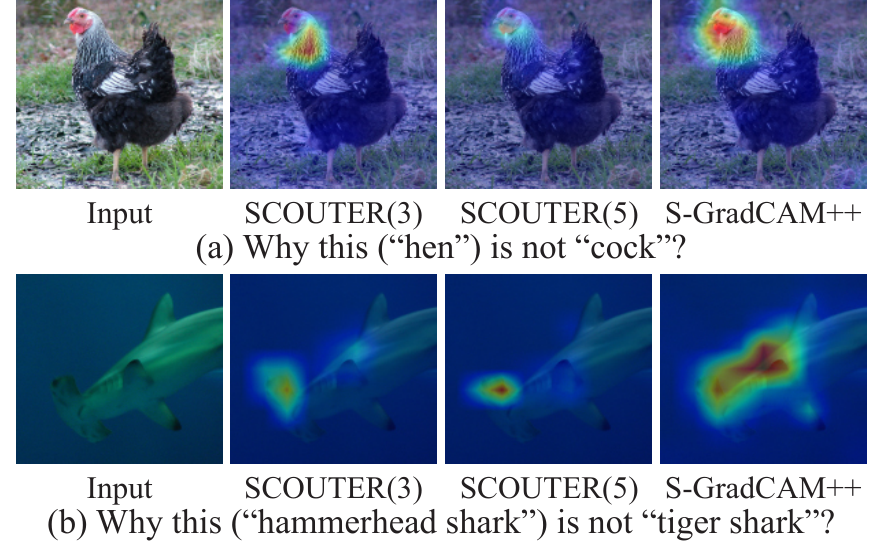}
	\caption{Visualized negative explanations using \SCOUTERnegative and an existing method. The numbers in the parentheses are the $\lambda$ values used in the SCOUTER training.}
	\label{fig_exp_area_size_examples_negative}
\end{figure}

\begin{figure*}[!t]
	\centering
	\includegraphics[width=\textwidth]{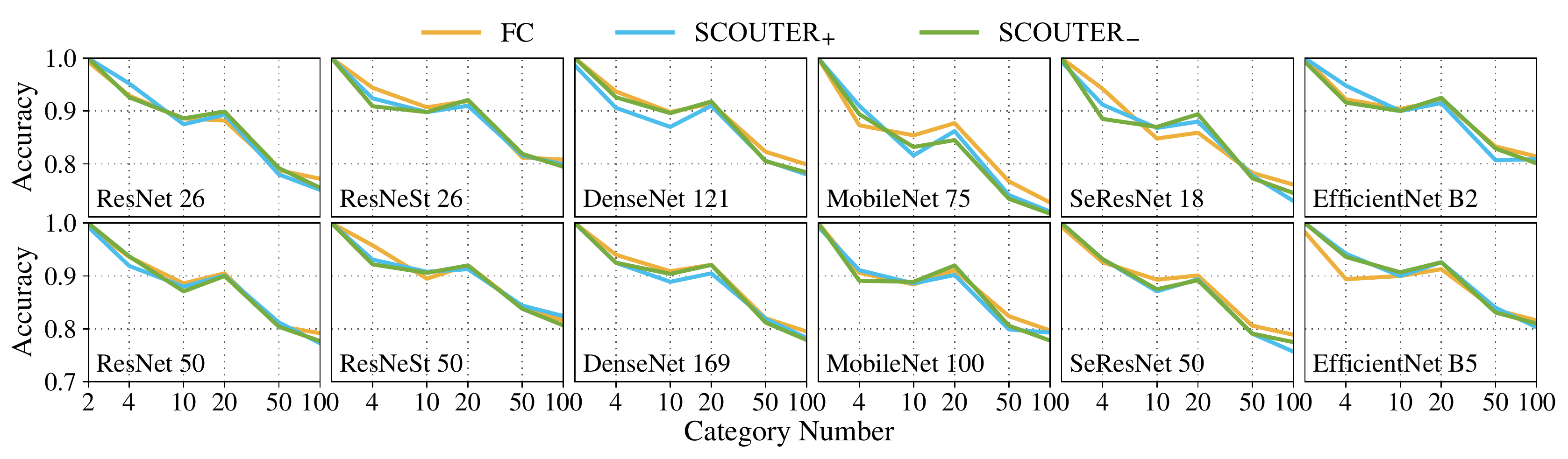}
	\caption{Classification performance of different models with FC classifier, \SCOUTERpositive ($\lambda=10$), and \SCOUTERnegative ($\lambda=10$). The horizontal axis is the category number $n$ (in the logarithmic scale), which is used to generate the training and test set with the first $n$ categories of ImageNet dataset; the vertical axis is the accuracy of the model.}
	\vspace{-0.1in}
	\label{fig_exp_accuracy}
\end{figure*}

Some visualization results are given in Figs.~\ref{fig_exp_area_size_examples} and \ref{fig_exp_area_size_examples_negative}. It can be seen that SCOUTER gives reasonable and accurate explanations. Comparing \SCOUTERpositive's explanation with \mbox{SS-CAM} \cite{Score-CAM}, and IBA \cite{IBA}, we find that \SCOUTERpositive can give explanations with more flexible shapes which fit the target objects better. For example, in the first row of Fig.~\ref{fig_exp_area_size_examples}, \SCOUTERpositive gives more accurate attention around the neck. In the second row, it accurately finds the individual entities. %With larger $\lambda$, it pinpoints small individual regions. % which are most responsible for the classification. % rather than covering inconsistent regions among entities. %In addition, in the last row, \SCOUTERpositive spots the dorsal fin, which is also an important clue to recognize sharks.
Compared with \mbox{SS-CAM}, IBA shows smaller explanatory regions. %, especially on the shark image.
However, IBA is less precise and less reasonable, %as is in the ostrich and chicken images, 
which is consistent with the numerical results in Table \ref{table_exp_explainability_measurement}.

In Fig.~\ref{fig_exp_area_size_examples_negative}, \SCOUTERnegative can also find the negative supports, \eg, the wattle of the hen, and the hammerhead and the fin of the shark.
In addition, although the negative variation of S-GradCAM++ performs well on the first row, its explanation in the second row does not well fit the object's shape and fails to pinpoint the key difference (the head).
%An ablation study is conducted in Section.~\ref{sec_ablation_study} to show the use of $\lambda$.
%More visualization results can be found in the supplementary material.

\subsection{Classification Performance}

%SCOUTER is a classifier that replaces FC classifiers; therefore, the classification performance is the main criterion. 
We compare SCOUTER and FC classifiers with several commonly used backbone networks with respect to the classification accuracy.
The results are summarized in Fig.~\ref{fig_exp_accuracy}. With the increase of the category number, both the FC classifier and SCOUTER show a performance drop. They show similar trends with respect to the category number.

\begin{figure}[!t]
\vspace{-0.1in}
	\centering
	\includegraphics[width=\columnwidth]{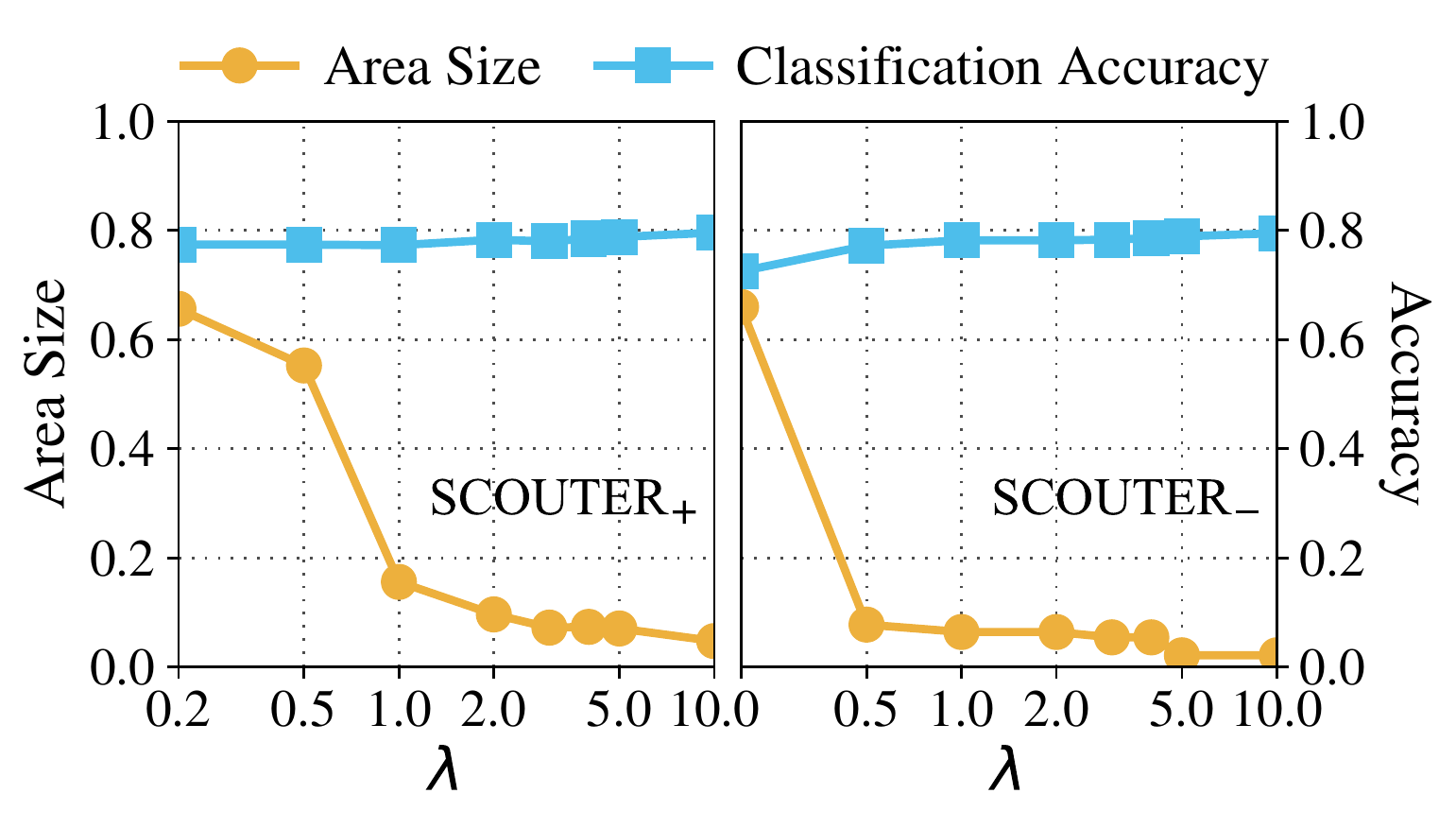}
	\caption{Relationships between $\lambda$ and explanation area sizes and between $\lambda$ and classification accuracy for the GT (\SCOUTERpositive, left) and LSC (\SCOUTERnegative, right) when $n=100$. The horizontal axis is in the logarithmic scale.}
	\label{fig_exp_lambda_changes}
\end{figure}

The relationship between $\lambda$, which controls the size of explanatory regions, and the classification accuracy is shown in Fig.~\ref{fig_exp_lambda_changes} for ResNeSt 26 model with $n = 100$. A clear pattern is that the area sizes of both \SCOUTERpositive and \SCOUTERnegative drop quickly with the increase of $\lambda$. %In addition, \SCOUTERnegative is more sensitive to the changes of $\lambda$. 
However, there is no significant trend in the classification accuracy, which should be because the cross-entropy loss term works well regardless of $\lambda$.

Also, according to the visualization results in Figs.~\ref{fig_exp_area_size_examples} and \ref{fig_exp_area_size_examples_negative}, a larger $\lambda$ does not simply decrease the explanatory regions' sizes. Instead, SCOUTER shifts its focus from some larger supports to fewer, smaller yet also decisive supports. For example, in the first row of Fig.~\ref{fig_exp_area_size_examples_negative}, when $\lambda$ is small, \SCOUTERnegative can easily make a decision that the input image is not a cock because of
unique feathers on the neck. With a larger $\lambda$, SCOUTER finds smaller combinations of supports (\ie, its wattle) and thus the explanation changes from the (larger) neck to the (smaller) wattle region.

%its heads and unique feathers on the neck. With a larger $\lambda$, SCOUTER finds smaller combinations of supports (\ie, its neck) and thus the explanation changes from the (larger) head region to the (smaller) neck region, and ultimately, to the (much smaller) wattle region.

We also summarize the classification performance of the FC classifier, \SCOUTERpositive ($\lambda=10$), and \SCOUTERnegative ($\lambda=10$) over ImageNet \cite{ImageNet}, Con-text \cite{Con-text}, and CUB-200 \cite{CUB-200} datasets in Table \ref{table_exp_classification_other_datasets}. The subsets with $n=100$ are adopted for ImageNet and CUB-200, while all 30 categories are used for the Con-text. The results show that SCOUTER can be generalized to different domains and has a comparable performance with the FC classifier over all datasets. %However, SCOUTER has some disadvantages ($4\%$ performance drop for \SCOUTERpositive and $2\%$ for \SCOUTERnegative) in the CUB-200 dataset that composed entirely of various species of birds, which reveals that it may be difficult to find enough discriminative \textit{supports} in highly-fine-grained tasks and it is easier to find negative \textit{supports} than positive \textit{supports} in these tasks. 
Also, SCOUTER's number of parameters is comparable %(smaller when $n>90$) 
to FC's (more details in supplementary material).

One drawback of SCOUTER is that its training is unstable when $n>100$. % there are more than $100$ categories. 
This is possibly because of the increasing difficulty in finding effective supports that consistently appear in all images of the same category but are not shared by other categories.  %More exploration is needed to improve its stability when dealing with a large number of categories. %Another limitation is that SCOUTER requires training to switch between positive and negative explanation, which may not be convenient for some applications.
This drawback limits the application of SCOUTER to small- or medium-sized datasets.

%\renewcommand*{\thefootnote}{\fnsymbol{footnote}}
%\footnotetext{A subset consisting of first $20$ categories.}
%\renewcommand*{\thefootnote}{\arabic{footnote}}

%\begin{table*}[!t]
%	\caption{P}
%	\label{table_exp_accuracy}
%	\centering
%	\begin{tabular}{cc}
%		\hline
%	    1&2\\
%		
%		\hline
%	\end{tabular}
%\end{table*}

\subsection{Case Study}\label{section_case_study}
\begin{figure}[!t]
	\centering
	\includegraphics[width=\columnwidth]{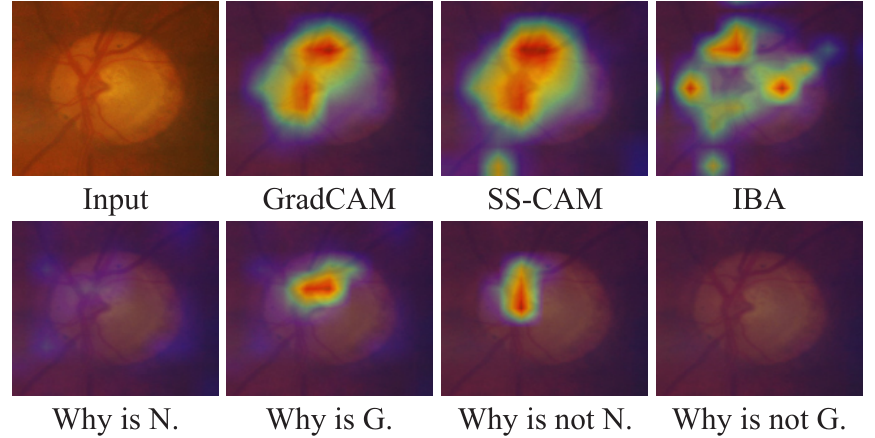}
	\caption{Explanations for a positive sample in the glaucoma diagnosis dataset. %Top row: input image and explanations by existing methods. Bottom row: explanations by \SCOUTERpositive (the first and second columns) and \SCOUTERnegative (the third and fourth columns) for normal (N.) and glaucoma (G.). 
	Bottom row is from SCOUTER +/- for normal (N.) and glaucoma (G.).}
	\label{fig_exp_case_study}
\end{figure}

SCOUTER uses the area loss, which constrains the size of support. This constraint can benefit some applications, including the classification of medical images, since small support regions can precisely show the symptoms and are more informative in some cases.
Also, there was no method that could give the negative explanation but it is actually needed (doctors need reasons to deny some diseases).
SCOUTER was designed upon these needs and is being tested in hospitals for glaucoma (Fig.~\ref{fig_exp_case_study}), artery hardening (supplementary material), \etc.

For glaucoma diagnosis, we tested SCOUTER with $\lambda=10$ over a publicly available dataset, \ie, ACRIMA \cite{ACRIMA}, which has two categories (normal and glaucoma). %The dataset size is $705$, in which $309$ samples are normal and $396$ samples are glaucoma. We split the dataset into train and test sets with a ratio of 7:3.
ResNeSt 26 is used as backbone. The results are shown in Table \ref{table_case_study}. We can see that both \SCOUTERpositive and \SCOUTERnegative get better performances than the FC classifier.
Besides, SCOUTER is preferred in this task as doctors are eager to know the precise regions in the optic disc that lead to the machine diagnosis. We can see that, in the visualization results in Fig.~\ref{fig_exp_case_study}, SCOUTER shows more precise and reasonable explanations that locate on some vessels in the optic disc and show clinical meanings (vessel shape change due to the enlarged optic cup), which are verified by doctors. %Comparing with the other XAI methods, 
Although IBA also gives small regions, they cover some unrelated or uninformative locations.
In addition, it is notable that the facts to admit category ``Glaucoma'' need not to match with the facts to deny ``Normal'', as they are only subsets of the support sets and an on-purpose negative explanation is especially helpful for the doctors when the machine decisions are against their expectations. 

%In the artery hardening diagnosis (Fig.~\ref{fig_artery_hardening}, more examples can be found in the supplementary material), the explanation should be precisely at crossing inside the white circle which shows vein shape change caused by the artery. We can see that existing methods cannot give precise regions locating in the symptom region and thresholding may not help.%lead to more inaccurate explanations.

\begin{table}[!t]
	\caption{Classification accuracy on various datasets.}
	\label{table_exp_classification_other_datasets}
	\centering
	\resizebox{0.95\columnwidth}{!}{%
    	\begin{tabular}{lccc}
    		\toprule
    	    
    	    Models  &  ImageNet & Con-text  & CUB-200\\
    	    \midrule
    	    ResNeSt 26 (FC) & \textbf{0.8080} & 0.6732 & \textbf{0.7538}\\
            ResNeSt 26 (\SCOUTERpositive) & 0.7991 & \textbf{0.6870} &  0.7362\\
            ResNeSt 26 (\SCOUTERnegative) & 0.7946 & 0.6866 & 0.7490 \\
            \midrule
    	    ResNeSt 50 (FC) & 0.8158 & 0.6918 & \textbf{0.7739}\\
            ResNeSt 50 (\SCOUTERpositive) & \textbf{0.8242} & \textbf{0.6943} & 0.7397\\
            ResNeSt 50 (\SCOUTERnegative) & 0.8066 & 0.6922 & 0.7600 \\
            \midrule
    	    ResNeSt 101 (FC) & 0.8255 & 0.7038 & \textbf{0.7804}\\
            ResNeSt 101 (\SCOUTERpositive) & 0.8251 & \textbf{0.7131} & 0.7428 \\
            ResNeSt 101 (\SCOUTERnegative) & \textbf{0.8267} & 0.7062 & 0.7643 \\
    		\bottomrule
    	\end{tabular}
	}
	\vspace{-0.1in}
\end{table}

%\begin{figure}[!t]
%	\centering
%	\includegraphics[width=\columnwidth]{LaTeX/figs/zfig_artery_hardening.pdf}
	
%	\caption{Explanations for a ``mild'' sample in the artery hardening diagnosis (4 classes). (b,c) are from SCOUTER +/-.}
%	\vspace{-0.2in}
%	\label{fig_artery_hardening}
%\end{figure}

\section{Conclusion}
An explainable classifier is proposed in this paper, with two variants, \ie, \SCOUTERpositive and \SCOUTERnegative, which can respectively give positive or negative explanation of the classification process. SCOUTER adopts an explainable variant of the slot attention, namely, xSlot attention, which is also based on the self-attention. Moreover, a loss is designed to control the size of explanatory regions. Experimental results prove that SCOUTER can give accurate explanations while keeping good classification performance.

\begin{table}[!t]
	\caption{Classification Performance on ACRIMA Dataset \cite{ACRIMA}.}
	\label{table_case_study}
	\centering
	\resizebox{0.95\columnwidth}{!}{%
    	\begin{tabular}{ccccccc}
    		\toprule
    	     Methods &  AUC & Acc. & Prec. & Rec. & F1 & Kappa\\
    	    \midrule
    	    FC & 0.9997 & 0.9857 & 0.9915 & 0.9831 & 0.9872 & 0.9710\\
    	    \SCOUTERpositive & \textbf{1.0000} & \textbf{1.0000} & \textbf{1.0000} & \textbf{1.0000} & \textbf{1.0000} & \textbf{1.0000}\\
    	    \SCOUTERnegative & 0.9999 & 0.9952 & \textbf{1.0000} & 0.9915 & 0.9957 & 0.9903\\
    		\bottomrule
    	\end{tabular}
	}
	\vspace{-0.15in}
\end{table}

\section{Acknowledgments}
This work was supported by Council for Science, Technology and Innovation (CSTI), cross-ministerial Strategic Innovation Promotion Program (SIP), ``Innovative AI Hospital System'' (Funding Agency: National Institute of Biomedical Innovation, Health and Nutrition (NIBIOHN)). This work was also supported by JSPS KAKENHI Grant Number 19K10662, 20K23343, 21K17764, MHLW AC Program Grant Number 20AC1006, and JST CREST Grant Number JPMJCR20D3, Japan.

%\section*{Ethical Impact} 
%Targeted research problem of XAI answers many questions in deep learning such as why and how in order to extend transparency. It can help in determining biasness of the model or data and it is highly informative and visual instead of a black box. 
%Moreover, the proposed method aims to control 

%A possible negative impact of this research is that the proposed classifier may introduce some bias based on the dataset (\eg, race, gender, ethnicity, \etc. in human datasets) while it should be unbiased irrespective of data. However, bias can be visualized using proposed attention visualization and may be dealt with proper modification in data or model.   

{\small
\bibliographystyle{ieee_fullname}
\bibliography{bib}
}
\vspace{6.7in}

%%%%%%%%%% Merge with supplemental materials %%%%%%%%%%
\pagebreak
\clearpage

\begin{strip}
\begin{center}
	 {\Large \bf SCOUTER: Slot Attention-based Classifier for Explainable Image Recognition\\(Supplementary Material) \par}
	
	 \vskip .5em
	 \vspace*{12pt}
\end{center}
\vspace{0.5 in}
\end{strip}
%%%%%%%%%% Merge with supplemental materials %%%%%%%%%%
%%%%%%%%%% Prefix a "S" to all equations, figures, tables and reset the counter %%%%%%%%%%
\setcounter{equation}{0}
\setcounter{figure}{0}
\setcounter{table}{0}
\setcounter{section}{0}
\setcounter{page}{1}
\makeatletter

\section{Computational Costs}

%\iftrue
%\clearpage
%\pagebreak

\begin{figure}[!t]
	\centering
	\includegraphics[width=\columnwidth]{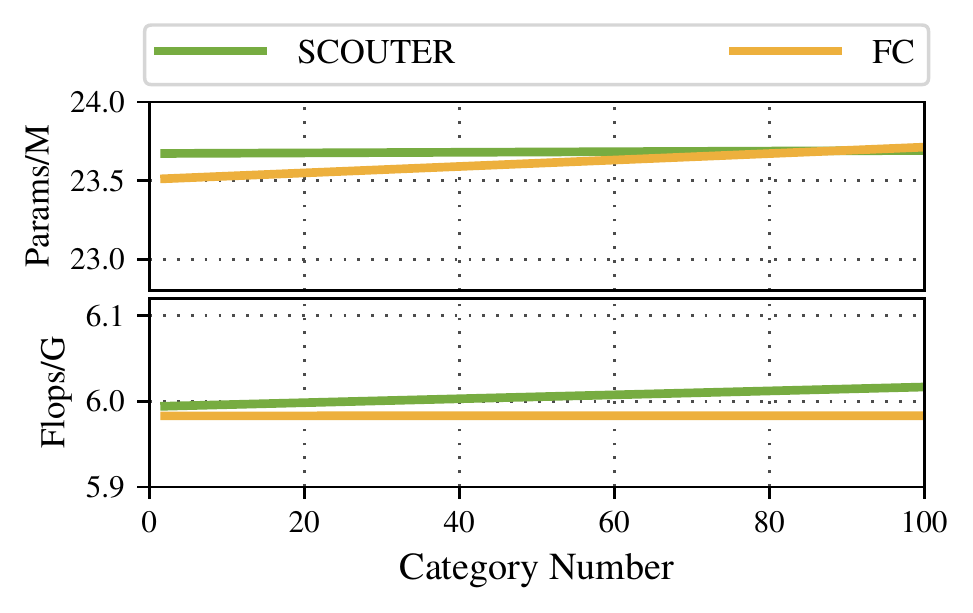}
	\caption{Flops and parameter sizes of SCOUTER and FC classifier with ResNet 50.}
	\label{fig_exp_params_flops}
\end{figure}

Table~\ref{table_exp_models} gives the cost comparison of SCOUTER and FC classifier. % with some commonly-used models. 
%The results in Table~\ref{table_exp_models} show that
We can see that, compared with the FC classifier, SCOUTER requires a similar computational cost (slightly higher) and a similar number of parameters (slightly lower). The increase in the computational cost (flops) is because the xSlot module has some small FC layers (\ie, $Q$ and $K$), GRU, and some matrix calculations. However, as shown in the lower part of Fig.~\ref{fig_exp_params_flops}, this is not very significant. % and it increases linearly.

On the other hand, as shown in the upper part of Fig.~\ref{fig_exp_params_flops}, SCOUTER has more parameters than the FC classifier when $n$ is roughly in $[0,90]$. This is because the FC layers and GRU, which are shared among all slots, have a certain number of parameters. For $n > 90$, SCOUTER uses fewer parameters than the FC classifier because there are only $c'$ ($64$ in our implementation) learnable parameters for each category. This is much less than the parameter size of the FC classifier, which usually needs much more parameters per class ($2,048$ parameters for ResNet 50).

Comparing to the differences in the computation costs and the numbers of parameters of different backbone models, the additional cost of SCOUTER is almost negligible.

\section{Components of xSlot Attention Module}
In SCOUTER, we adopt a variant of the slot attention \cite{Slot}. We make some essential modifications to several components in order to enable explainable classification, while other components, \ie the gated recurrent unit (GRU) and position embedding (PE), remain unchanged, whose effects on the classification as well as the explainability are unexplored. To test the performance of the SCOUTER with and without these components, we consider two variants of SCOUTER. The first one is the SCOUTER without GRU, in which we replace the GRU component, which is used to update slot weights, with an average operation. The second variant is the SCOUTER without PE, where flattened input features are directly used without adding position information. 

In Table~\ref{table_exp_ablation_study}, we show the performances of \SCOUTERpositive and \SCOUTERnegative as well as their variants in several performance metrics including computation costs, classification accuracy, and explainability. We can see that SCOUTER with all the components gets better results in most metrics than the variants, except for computation costs. The absence of GRU or PE not only causes a decrease of the classification accuracy, but also some deterioration on all explainability metrics, which proves their necessity.

\begin{table}[!t]
	\caption{Cost comparison of SCOUTER and FC classifier ($n=100$ and input images are with the size of $260\times260$).}
	\label{table_exp_models}
	\centering
	\resizebox{\columnwidth}{!}{%
    	\begin{tabular}{crrcrr}
    		\toprule
    	    \multirow{2}{*}{Models} & \multicolumn{2}{c}{Params (M)} && \multicolumn{2}{c}{Flops (G)}\\
    	    \cline{2-3}\cline{5-6}
    	    & FC$\quad$ & SCOUTER$\ $ && $\ $FC$\quad$ & SCOUTER\\
    	    \midrule
    	    ResNet 26 \cite{ResNet} &   14.1511 & \textbf{14.1298}$\quad$ && \textbf{3.4238} & 3.4565$\quad$\\
    	    ResNet 50 \cite{ResNet} &  23.7129 &  \textbf{23.6916}$\quad$ && \textbf{5.9830} & 6.0171$\quad$\\
    	    ResNeSt 26 \cite{ResNeSt} & 15.2253 & \textbf{15.2041}$\quad$ && \textbf{5.1803} & 5.2130$\quad$\\
    	    ResNeSt 50 \cite{ResNeSt} & 25.6391 & \textbf{25.6179}$\quad$ && \textbf{7.7430} & 7.7762$\quad$\\
    	    DenseNet 121 \cite{DenseNet} & 7.0564 & \textbf{7.0719}$\quad$ && \textbf{3.7536} & 3.7805$\quad$ \\
    	    DenseNet 169 \cite{DenseNet} & 12.6510 &\textbf{12.6435}$\quad$&& \textbf{4.4396} & 4.4683$\quad$\\
    	    MobileNet 75 \cite{MobileNet} & 1.1194 &\textbf{0.6537}$\quad$&& \textbf{0.0563} & 0.0812$\quad$\\
    	    MobileNet 100 \cite{MobileNet} & 4.3301 &\textbf{3.0859}$\quad$&& \textbf{0.3154} & 0.3421$\quad$\\
    	    SeResNet 18 \cite{SeNet} & 11.3169 &\textbf{11.3509}$\quad$&& \textbf{2.6473} & 2.6726$\quad$\\
    	    SeResNet 50 \cite{SeNet} & 26.2439 &\textbf{26.2226}$\quad$&& \textbf{5.6758} & 5.7098$\quad$\\
    	    EfficientNet B2 \cite{Efficientnet} & 7.8419 &\textbf{7.8437}$\quad$&& \textbf{1.0250} & 1.0564$\quad$\\
    	    EfficientNet B5 \cite{Efficientnet} & 28.5457 &\textbf{28.5244}$\quad$&& \textbf{3.6391} & 3.6721$\quad$\\
    		
    		\bottomrule
    	\end{tabular}
	}
\end{table}

\begin{table*}[!t]
	\caption{Performance comparison of SCOUTER and its variants on a subset ($n=100$) of the ImageNet dataset. $\lambda$ is set to $10$ during training and ResNeSt 26 is adopted as the backbone. The explanation performance is measured on the GT category for the positive explanation and on the least similar class (LSC) for the negative explanation.}
	\label{table_exp_ablation_study}
	\centering
	\resizebox{.8\textwidth}{!}{%
    	\begin{tabular}{c|c|cc|c|ccc}
    		\toprule
    		\multirow{2}{*}{Explanation Type} & \multirow{2}{*}{Variants} & \multicolumn{2}{c|}{Computational Costs} & Classification & \multicolumn{3}{c}{Explainability}\\
    	    & & Params (M) & Flops (G) & Accuracy & Precision & IAUC & DAUC\\
    	    \midrule
    	    \multirow{3}{*}{Positive}& \SCOUTERpositive & 15.2041 & 5.2130 & \textbf{0.7991} & \textbf{0.9257} & \textbf{0.7647} & \textbf{0.2713}\\
    	    & \textit{w.o.} GRU & \textbf{15.1791} & \textbf{5.1901} & 0.7961 & 0.9219 & 0.7456 & 0.2866\\
    	    & \textit{w.o.} PE & 15.2041 & 5.2130 & 0.7974 & 0.8973 & 0.7557 & 0.3002\\
    	    %& \textit{w.} CNN & 37.4109 & 6.9928 \\
    	    \midrule
    	    \multirow{3}{*}{Negative}& \SCOUTERnegative & 15.2041 & 5.2130 & \textbf{0.7946} & \textbf{0.8101} & \textbf{0.6730} & \textbf{0.7333}\\
    	    & \textit{w.o.} GRU & \textbf{15.1791} & \textbf{5.1901} & 0.7910 & 0.7904 & 0.5959 & 0.7529\\
    	    & \textit{w.o.} PE & 15.2041 & 5.2130 & 0.7903 & 0.8067 & 0.6141 & 0.7661\\
    	    %& \textit{w.} CNN & 37.4109 & 6.9928 \\
    		
    		\bottomrule
    	\end{tabular}
	}
\end{table*}

\section{Classification Performance when $n>100$}

\begin{figure}[!t]
	\centering
	\includegraphics[width=\columnwidth]{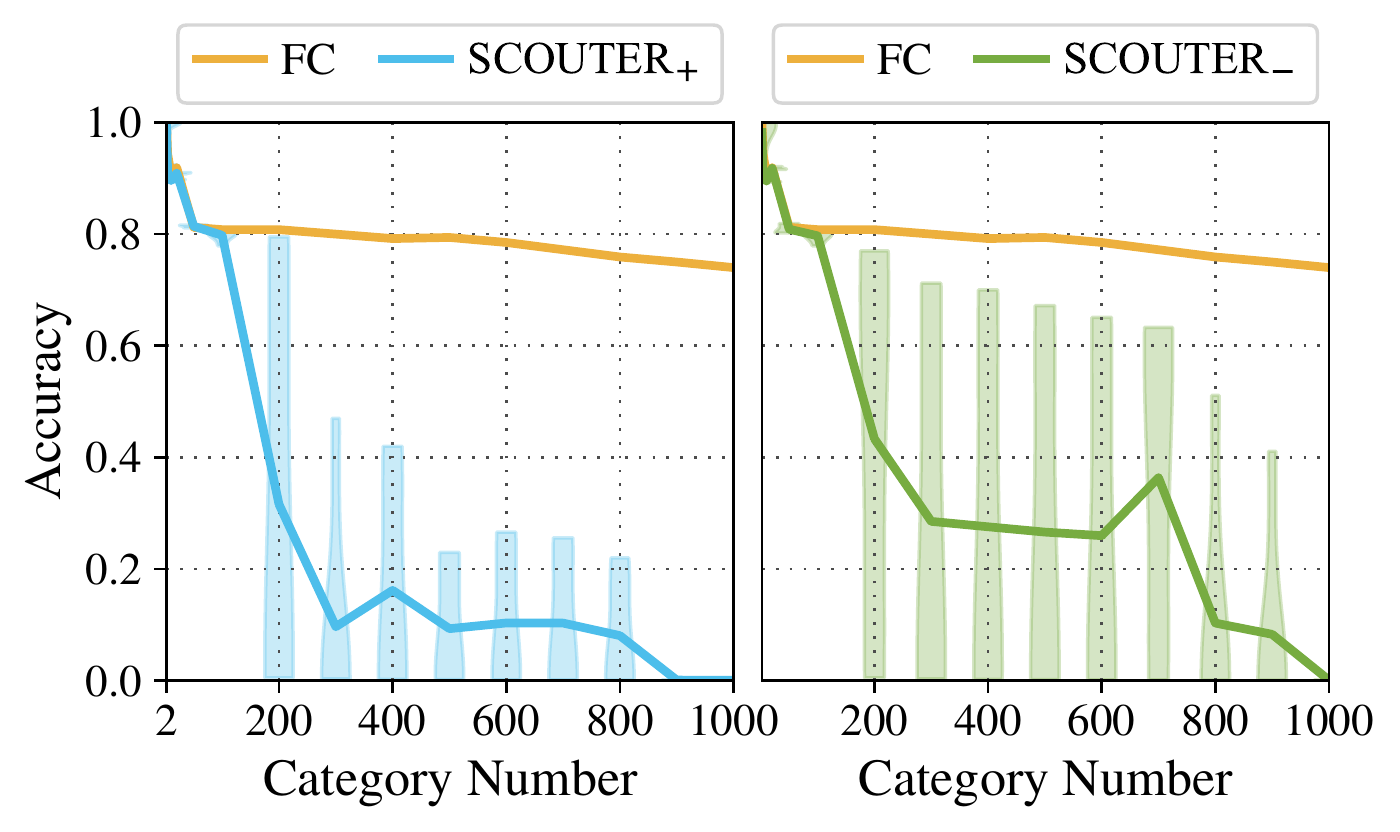}
	\caption{The classification performance of FC classifier, \SCOUTERpositive, and \SCOUTERnegative when $2\leq n\leq1000$. We show the violin plots as well as the average value for \SCOUTERpositive and \SCOUTERnegative, while the FC classifier is only with the average value.}
	%\vspace{-0.15in}
	\label{fig_exp_large_dataset}
\end{figure}

Training of SCOUTER becomes unstable when the category number $n$ of the ImageNet \cite{ImageNet} subsets is larger than $100$. One possible reason is that it is difficult to find consistent and discriminative supports when there are many categories. Fig.~\ref{fig_exp_large_dataset} shows the classification performance when $n>100$. The number of independent runs of training is increased to $5$ as the training process becomes unstable and often results in failures (low classification accuracy) when $n>100$. $\lambda$ is set to $10$. ResNeSt 26 \cite{ResNeSt} is adopted as the backbone, with batch size of $70$ and training epoch number of $20$ (both are same as the settings of the experiments in the main paper). We can see that, although sometimes \SCOUTERpositive and \SCOUTERnegative can achieve similar performance with the FC classifier when $n<400$, they become significantly unstable with the increase of category number $n$.
As stated in the main paper, SCOUTER can only be used in small-or medium-sized datasets due to this issue.

\section{Inter-and Intra-Category Explanation}
\begin{figure*}[p]
	\centering
	\includegraphics[width=1\textwidth]{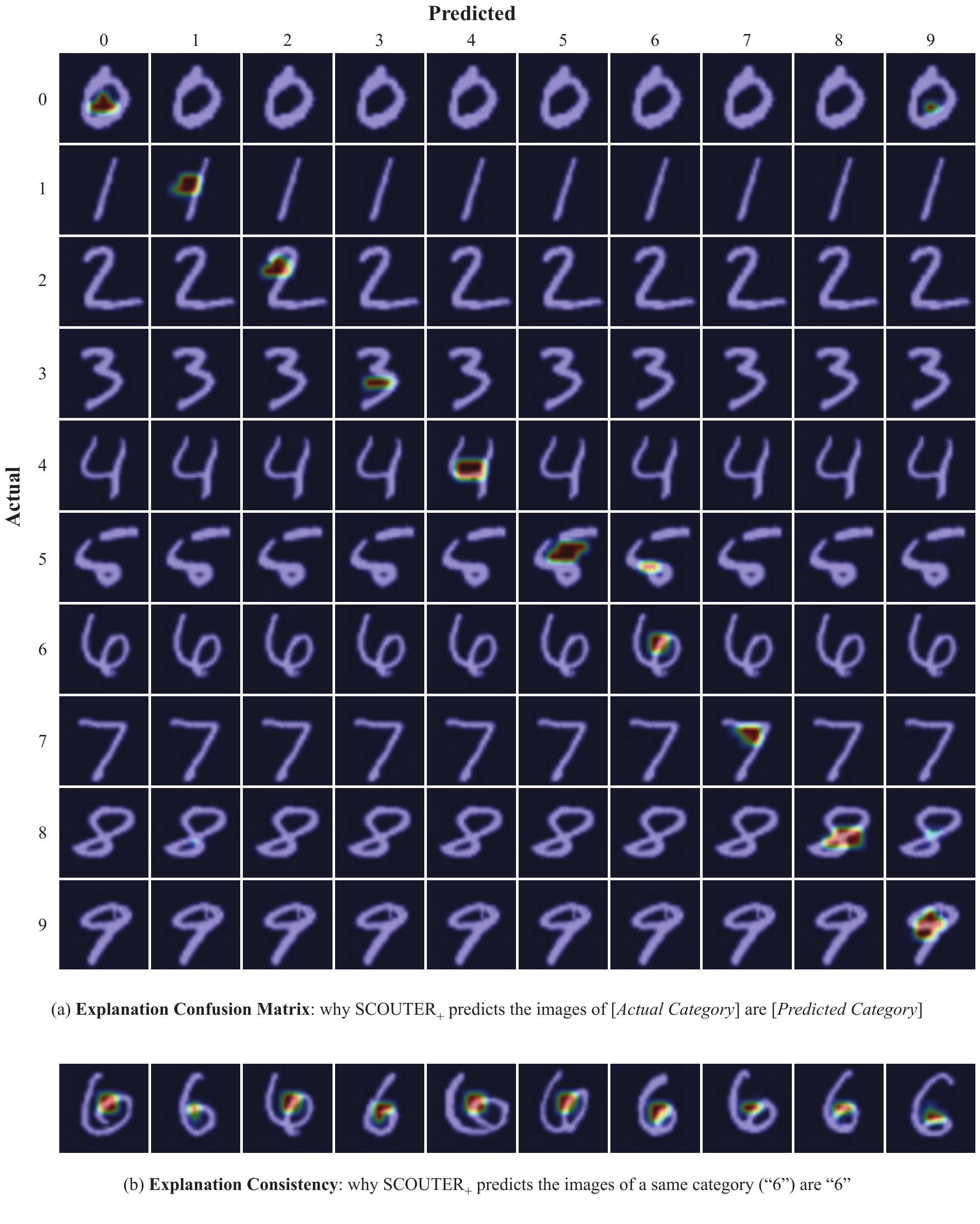}
	\caption{Visualized positive explanations using \SCOUTERpositive (with ResNet 18 \cite{ResNet} and $\lambda=1$) on the MNIST dataset \cite{MNIST}.}
	\label{fig_exp_confusion_matrix1}
\end{figure*}

\begin{figure*}[p]
	\centering
	\includegraphics[width=1\textwidth]{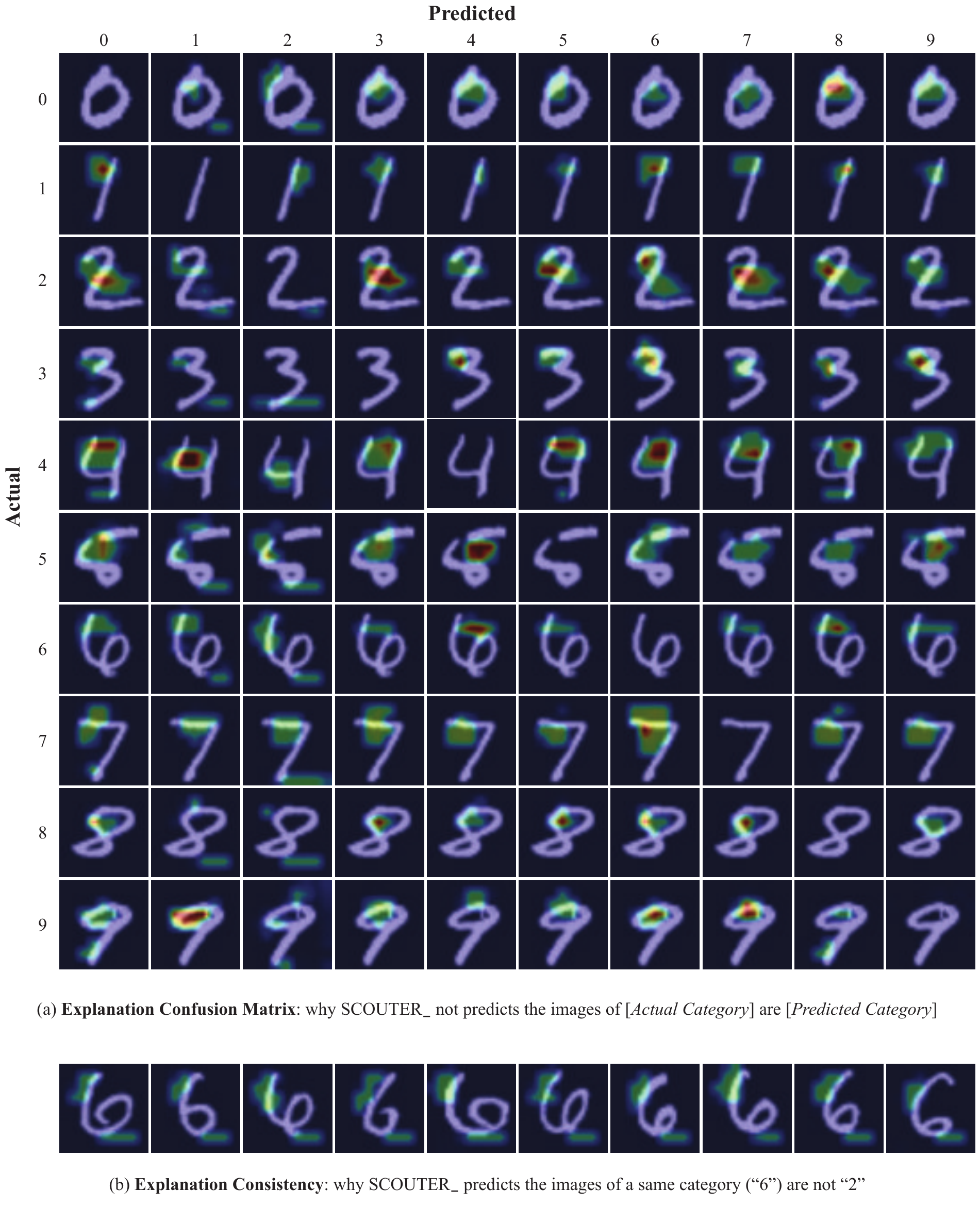}
	\caption{Visualized negative explanations using \SCOUTERnegative (with ResNet 18 \cite{ResNet} and $\lambda=1$) on the MNIST dataset \cite{MNIST}.}
	\label{fig_exp_confusion_matrix2}
\end{figure*}

To better understand (i) what supports SCOUTER uses as the basis for its decision making, (ii) how these supports can be differentiated among different categories, and (iii) whether they are being consistent for images in the same category, we give some additional visualization on the MNIST dataset \cite{MNIST} in Figs.~\ref{fig_exp_confusion_matrix1} and \ref{fig_exp_confusion_matrix2} for \SCOUTERpositive and \SCOUTERnegative, respectively. %MNIST is adopted here because we are all familiar with numerical characters and we can well understand the similarities and dissimilarities among categories.
MNIST is adopted here as similarities and dissimilarities among categories (digits) are obvious and are easier to understand than ImageNet. In these two figures, (a) is for the inter-category visualization, which shows what the supports for the ``Predicted'' category look like given the image of the ``Actual'' category. Whereas, (b) is for intra-category visualization, which shows the support for different images of the same category. For the latter, we use the digit \texttt{6} as an example and the first ten samples of category \texttt{6} in the test set of MNIST are used.

In the inter-category visualization in Fig.~\ref{fig_exp_confusion_matrix1}, we can see that \SCOUTERpositive successfully finds supports for the images of ground-truth (GT) categories. Notably, it also finds weaker supports for some categories with similar appearances, \eg, the supports for the prediction of ``why \texttt{5} is \texttt{6}'' (as the lower half of this hand-wrote \texttt{5} digit is a little confusing and is very close to the lower part of \texttt{6}), as well as the prediction of ``why \texttt{0} is \texttt{9}'' and ``why \texttt{8} is \texttt{9}'' (both \texttt{0} and \texttt{8} have a circle like the one in \texttt{9}).

Similarly, in Fig.~\ref{fig_exp_confusion_matrix2}, we can see that \SCOUTERnegative finds no supports for the images of the GT categories, while it finds strong supports for the non-GT categories. As digit recognition is an easy task, \SCOUTERnegative can use some very simple supports to deny most non-GT categories. For example, in the prediction of ``why \texttt{1} is not [non-GT categories]'', all the slots of \SCOUTERnegative find that the top end of the vertical stroke is \texttt{1}'s unique pattern, thus, they can deny all other categories with this support. Among some visually similar categories, the negative explanations are more informative. For example, in the visualization of ``why \texttt{9} is not \texttt{1}'' and ``why \texttt{9} is not \texttt{7}'', \SCOUTERnegative precisely highlights the discriminative regions, without which \texttt{9} will look like the other two digits.

Also, in intra-category visualization, both \SCOUTERpositive and \SCOUTERnegative show consistent supports for the images of the same category. When predicting ``why \texttt{6} is \texttt{6}'', \SCOUTERpositive always looks at the region close to the crossing point of the bottom circle and vertical stroke. For explanation ``why \texttt{6} is not \texttt{2}'', \SCOUTERnegative always recognizes the presence of vertical stroke, which does not exist in the digit \texttt{2}, as well as the missing of the bottom horizon stroke, which is essential for \texttt{2}.

%\vspace{-0.1in}
\section{Some More Visualizations}
In this section, we show more visualization results for ImageNet using SCOUTER and competing methods, including I-GOS \cite{IGOS}, IBA \cite{IBA}, CAM \cite{CAM}, GradCAM \cite{Grad-CAM}, GradCAM++ \cite{GradCAM++}, S-GradCAM++ \cite{Smooth_GradCAM++}, Score-CAM \cite{Score-CAM}, SS-CAM \cite{SS-CAM}, and Extremal Perturbation \cite{Perturbation_understanding}.

\begin{figure*}[p]
	\centering
	\includegraphics[width=1\textwidth]{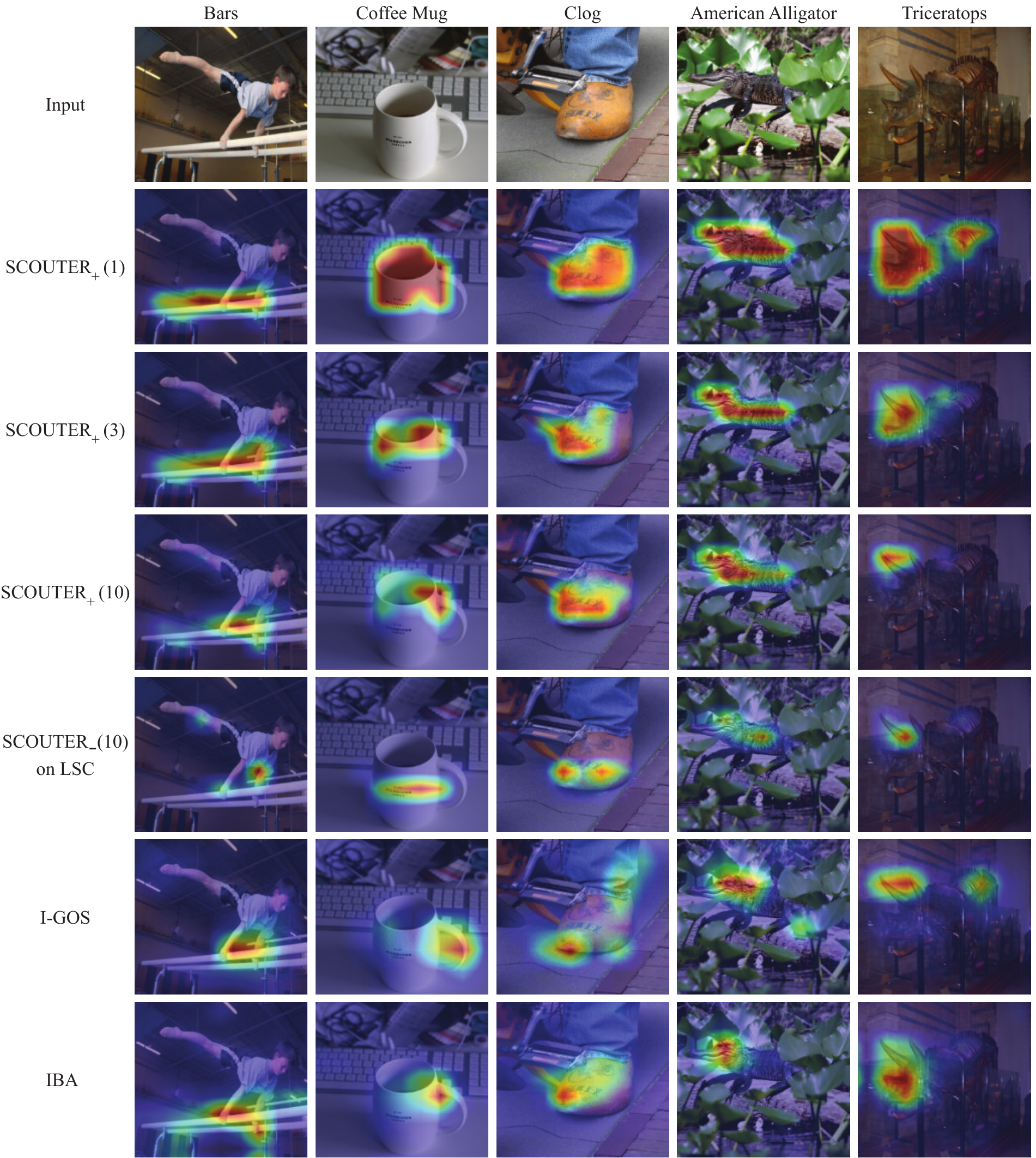}
	\caption{More examples of visualized positive explanations (Part 1). The number in parentheses represents the $\lambda$ value used in the SCOUTER training.}
	\label{fig_exp_confusion_more_visualization_results_positive1}
\end{figure*}

\begin{figure*}[p]
	\centering
	\includegraphics[width=1\textwidth]{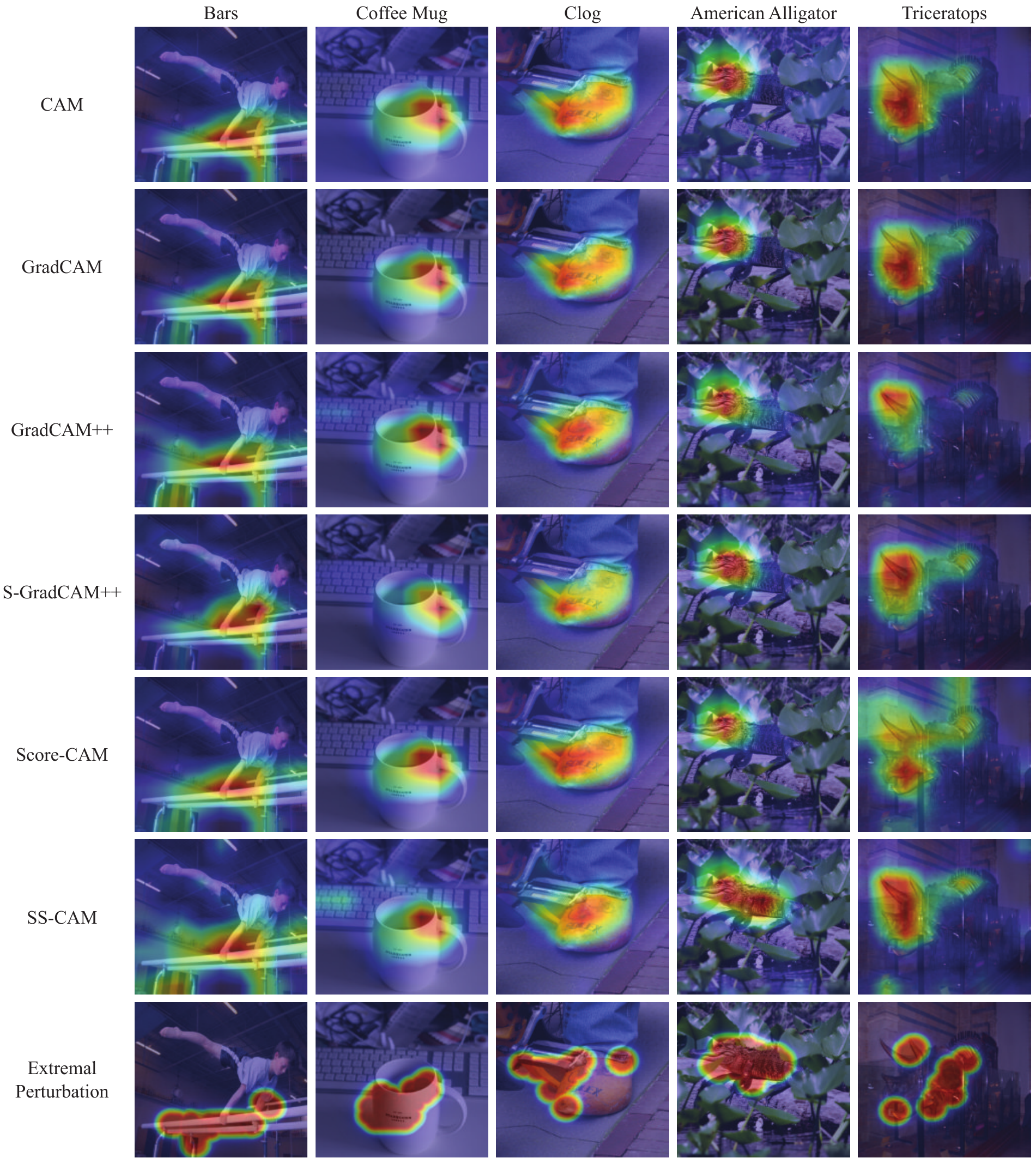}
	\caption{More examples of visualized positive explanations (Part 2). The number in parentheses represents the $\lambda$ value used in the SCOUTER training.}
	\label{fig_exp_confusion_more_visualization_results_positive2}
\end{figure*}

\begin{figure*}[p]
	\centering
	\includegraphics[width=.95\textwidth]{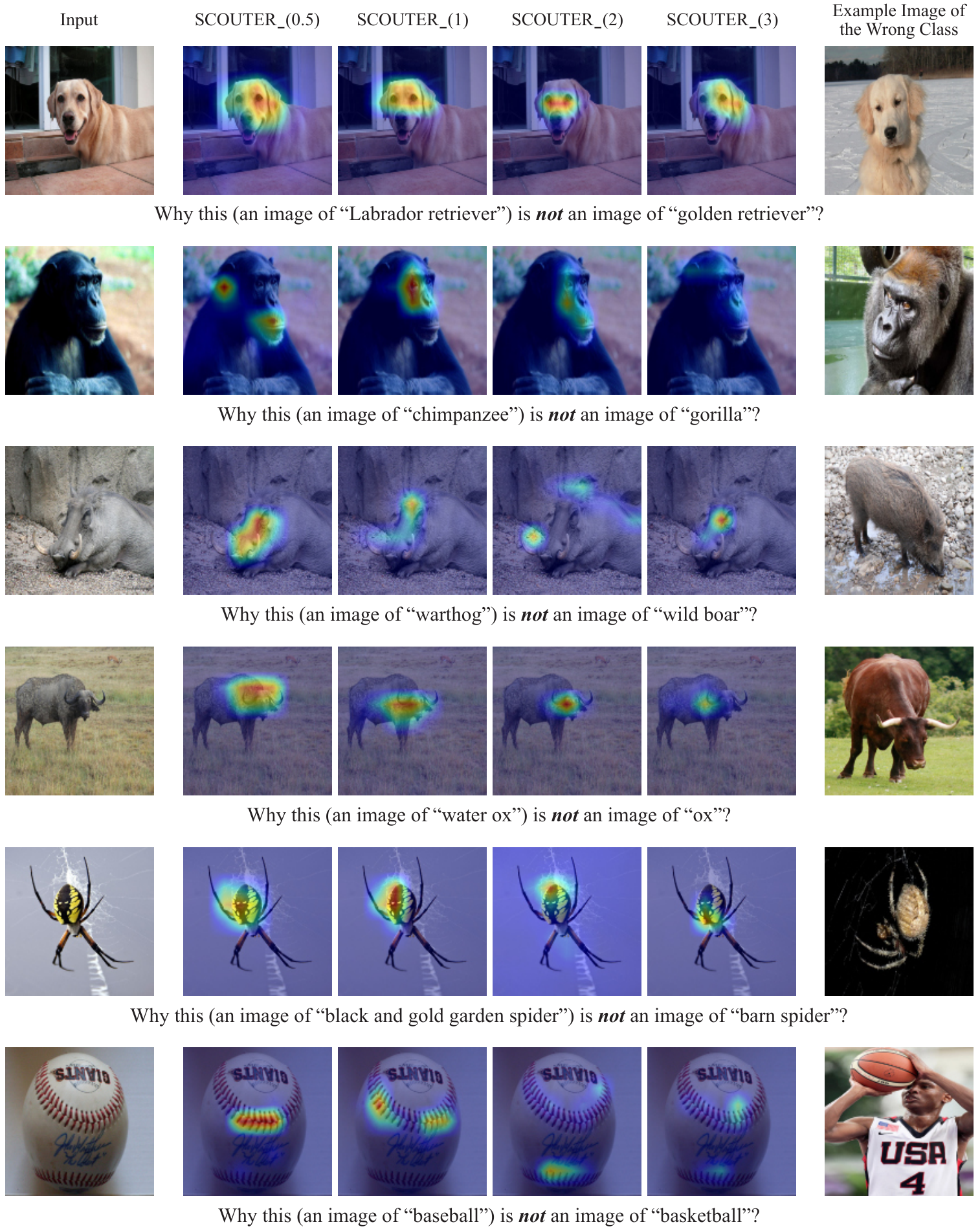}
	\caption{More examples of visualized negative explanations for similar categories. The number in parentheses represents the $\lambda$ value used in the SCOUTER training.}
	\label{fig_exp_confusion_more_visualization_results_negative}
\end{figure*}

Subsets with $n=100$ categories are used for training and visualization. Besides the first $n$ categories (as used in the main paper), we also use several other subsets (with the same category number) in the ImageNet dataset, in order to provide visualizations with more diversity. Figs.~\ref{fig_exp_confusion_more_visualization_results_positive1} and \ref{fig_exp_confusion_more_visualization_results_positive2} give examples of the positive explanation, while Fig.~\ref{fig_exp_confusion_more_visualization_results_negative} gives examples of the negative explanation.

Among the positive explanations, we can see that \SCOUTERpositive can find reasonable and precise supports. Especially for the image of ``\texttt{parallel bars}'', \SCOUTERpositive can provide an explanatory region along the horizon bar. In addition, \SCOUTERnegative with the least similar class (LSC) also finds supports on the foreground objects, which can be used to deny the LSC categories but are not enough for admitting the GT category, which conforms the quantitative results in the main paper. 

Moreover, as shown in Fig.~\ref{fig_exp_confusion_more_visualization_results_negative}, \SCOUTERnegative can give very detailed explanations when different categories with high visual similarities, \eg, the differences in the eyes and ears between ``\texttt{Labrador retriever}'' and ``\texttt{golden retriever}'', and the differences of the horn between ``\texttt{water ox}'' and ``\texttt{ox}''. %This kind of information is very useful for certain applications.

\begin{figure*}[p]
	\centering
	\includegraphics[width=.95\textwidth]{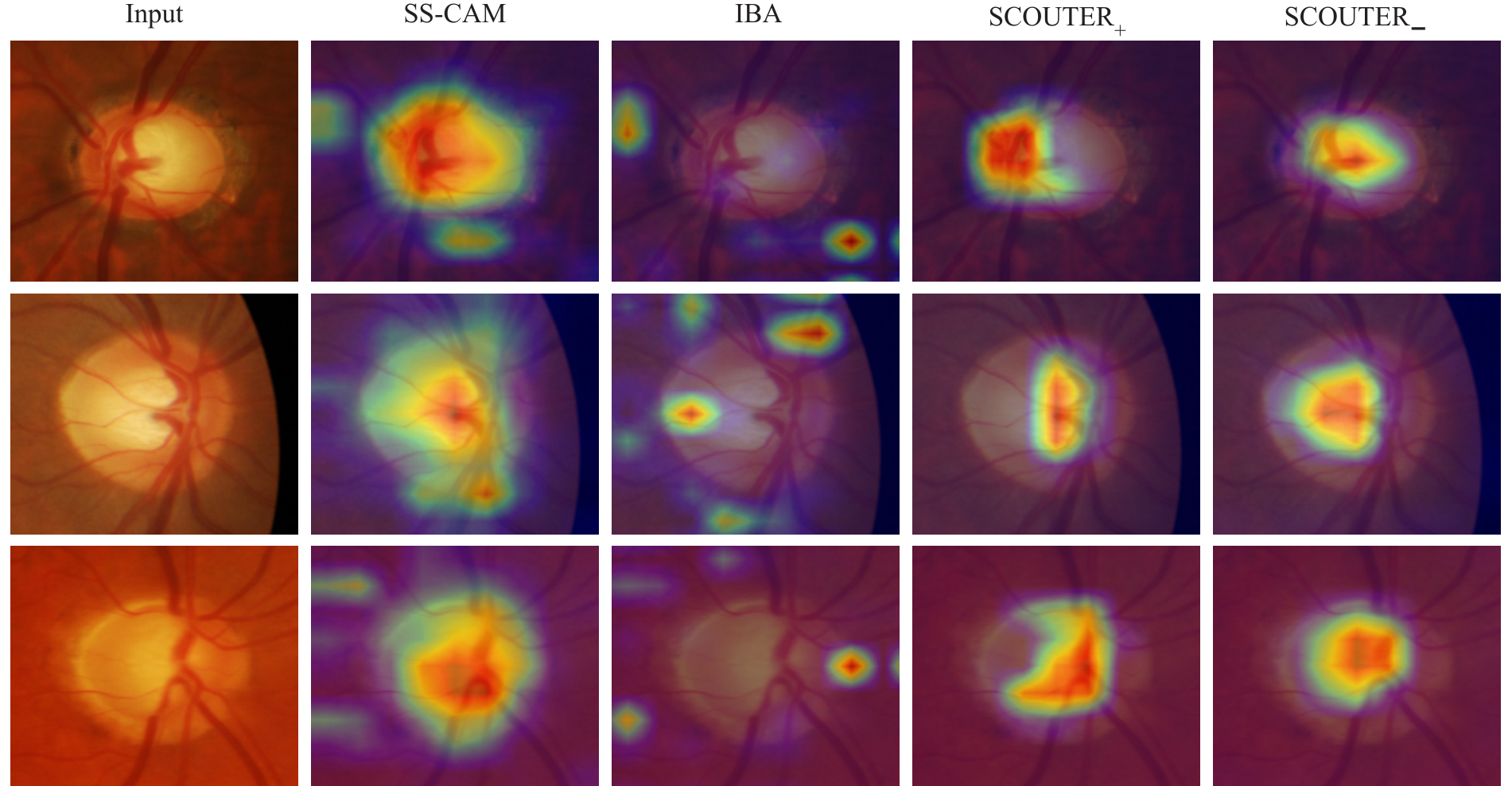}
	\caption{More examples of visualized explanations for glaucoma diagnosis on three positive samples using SS-CAM, IBA, \SCOUTERpositive, and \SCOUTERnegative. The first three methods are for ``why this is \texttt{glaucoma}'' while \SCOUTERnegative is for ``why this is not \texttt{normal}''. SCOUTER shows explanations covering only related regions (vessel shape changes), which have been validated by two doctors.}
	\label{fig_exp_area_size_examples_supp_glaucoma}
\end{figure*}

\begin{figure*}[p]
	\centering
	\includegraphics[width=.95\textwidth]{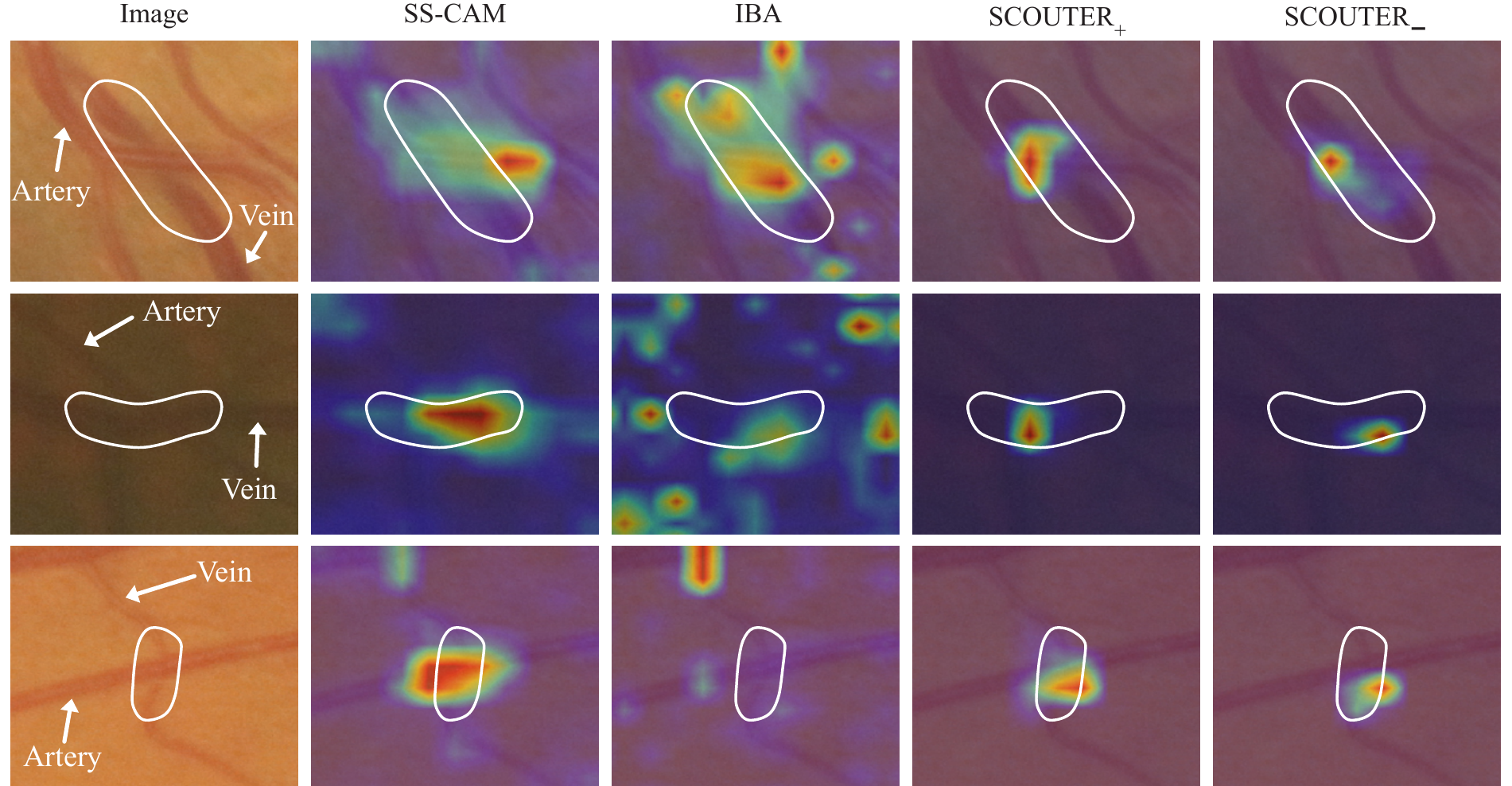}
	\caption{More examples of visualized explanations for artery hardening diagnosis on three ``\texttt{moderate}'' samples using SS-CAM, IBA, \SCOUTERpositive, and \SCOUTERnegative. The first three methods are for ``why this is \texttt{moderate}'' while \SCOUTERnegative is for ``why this is not \texttt{none}''. The white circles give the approximate location of the symptoms (shape changes on the vessel wall of the vein, which are caused by the increased blood pressure in the artery). SCOUTER gives precise explanations which are mostly within the symptom region and precisely on the wall of the vein. The explanations of SS-CAM are off the target in the first row and on the wrong vessels (artery) in the first and the third rows while IBA fails to find the symptom in the second and the third rows.}
	\label{fig_exp_area_size_examples_supp_artery_hardening}
\end{figure*}

Figs.~\ref{fig_exp_area_size_examples_supp_glaucoma} and \ref{fig_exp_area_size_examples_supp_artery_hardening} show some more examples of two medical applications (glaucoma diagnosis and artery hardening diagnosis). We can see that \SCOUTERpositive and \SCOUTERnegative perform well in both tasks. 
%In the glaucoma diagnosis task, SCOUTER shows explanations covering only related regions (vessel shape changes). In the artery hardening task, SCOUTER gives precise explanations which are mostly within the symptom region. As a comparison, the explanations of SS-CAM is off target in row 1 and on the wrong vessels (artery) in rows 1\&3 while IBA fails to find the symptom in rows 2\&3.

\end{document}